\documentclass{article}

\usepackage{PRIMEarxiv}

\usepackage[utf8]{inputenc} % allow utf-8 input
\usepackage[T1]{fontenc}    % use 8-bit T1 fonts
\usepackage{hyperref}       % hyperlinks
\usepackage{url}            % simple URL typesetting
\usepackage{booktabs}       % professional-quality tables
\usepackage{amsfonts}       % blackboard math symbols
\usepackage{nicefrac}       % compact symbols for 1/2, etc.
\usepackage{microtype}      % microtypography
\usepackage{lipsum}
\usepackage{fancyhdr}       % header
\usepackage{graphicx}       % graphics
\graphicspath{{media/}}     % organize your images and other figures under media/ folder
\usepackage{amsthm}
\usepackage{algorithm}
\usepackage{algorithmic}
\usepackage[switch]{lineno}
\usepackage{textcomp}
\usepackage{amsmath,amssymb,amsfonts}
\usepackage{enumitem}
\usepackage{caption}
\usepackage{subcaption}
\usepackage{multirow} 
\usepackage{xcolor}

\newtheorem{assumption}{\rm \bfseries{ Assumption}}
\newtheorem{problem}{\rm \bfseries{Problem}}
\newtheorem{theorem}{\rm \bfseries{Theorem}}
\newcommand{\sysname}{GLIDER}

\title{Graphs Generalization under Distribution Shifts
}

\author{
  Qin Tian \\
  Tianjin University \\
  Tianjin, China\\
  \texttt{tianqin123@tju.edu.cn} \\
    \And
    Wenjun Wang \\
  Tianjin University \\
  Tianjin, China\\
  \texttt{wjwang@tju.edu.cn} \\
  \And
  Chen Zhao\thanks{This work is supervised by Chen Zhao.}  \\
  Baylor University \\
  Waco, Texas, USA\\
  \texttt{chen\_zhao@baylor.edu} \\
  \And
  Minglai Shao\thanks{Corresponding author} \\
  Tianjin University \\
  Tianjin, China\\
  \texttt{shaoml@tju.edu.cn}\\
  \And
  Wang Zhang \\
  Tianjin University \\
  Tianjin, China\\
  \texttt{wangzhang@tju.edu.cn}\\
  \And
  Dong Li \\
  Tianjin University \\
  Tianjin, China\\
  \texttt{ld2022244154@tju.edu.cn} \\
}

\begin{document}
\maketitle

\begin{abstract}
Traditional machine learning methods heavily rely on the independent and identically distribution (\textit{i.i.d}) assumption, which imposes limitations when the test distribution deviates from the training distribution.
To address this crucial issue, out-of-distribution (OOD) generalization, which aims to achieve satisfactory generalization performance when faced with unknown distribution shifts, has made a significant process.
However, the OOD method for graph-structured data currently lacks clarity and remains relatively unexplored due to two primary challenges. Firstly, distribution shifts on graphs often occur simultaneously on node attributes and graph topology. Secondly, capturing invariant information amidst diverse distribution shifts proves to be a formidable challenge.
To overcome these obstacles, in this paper, we introduce a novel framework, namely Graph Learning Invariant Domain genERation (\sysname{}).
The goal of it is to (1) diversify variations across domains by modeling the potential seen or unseen variations of attribute distribution and topological structure, and (2) minimize the discrepancy of the variation in a representation space where the target is to predict semantic labels.
Technically, \sysname{} consists of three stages.
In the first stage, a graph ($G'$) is generated from the original graph ($G$) by only changing the distribution of node attributes.
Then, \sysname{} generates multiple graphs ($G''$) in synthetic domains from $G'$, wherein node attributes stay the same but topological structure changes in each $G''$.
Finally, the framework minimizes the semantic gap for each class between $G$ and $G''$ by learning a domain-invariant classifier.
We validate the effectiveness of \sysname{} by conducting empirical studies on four real-world graph datasets.
Extensive experiment results indicate that our model outperforms baseline methods on node-level OOD generalization across domains in distribution shift on node features and topological structures simultaneously.
\end{abstract}

% keywords can be removed
% \keywords{First keyword \and Second keyword \and More}

\section{Introduction}
\label{sec:intro}
In recent years, out-of-distribution (OOD) generalization has gained prominence as an approach to enhance the performance of machine learning methods when confronted with unknown distributions~\cite {muandet2013domain}.
The goal of OOD generalization is to find a predictor that is trained using data drawn from a family of related training domains and then evaluated on a distinct and unseen test domain~\cite{robey2021model}.
Indeed, numerous studies have demonstrated the effectiveness of OOD generalization methods on Euclidean data, such as images or molecules that satisfy the \textit{i.i.d} assumption. 
However, graph-structured data, prevalent in real-world applications, has garnered comparatively less attention despite its substantial importance in high-stakes graph applications. 
These applications include but are not limited to, molecule prediction~\cite{hu2020open}, financial analysis~\cite{yang2021financial}, and criminal justice~\cite{agarwal2021towards}.
% \textbf{The Case for OOD Generalization on Graphs:}
% The presence of distribution shifts in graph-structured data is widespread in real-world scenarios.
For example, social networks consist of users (nodes) who interact with each other and influence each other's behaviors (labels).
The structure of these interactions highly depends on when and where data is collected.
% Likewise, as the web page network described in Fig.~\ref{fig:instance}, nodes represent web pages, and edges are hyperlinks between them. 
% The attribute distribution (bag-of-words) and topological structure (hyperlinks) of web pages have the potential to change over time. 
The goal of OOD generation on graphs is to achieve stable and accurate prediction when distribution shifts occur on graph-structured data with complex intrinsic connections.

% \textbf{Challenges:} 
Due to the incorporation of distribution variations with diverse characteristics in graph-structured data, two main challenges arise: (1) Distribution shifts on the graph can manifest simultaneously in both node attributes and topology, resulting in a substantial and intricate variation. (2) Consequently, the variation poses challenges in capturing invariant information across different domains.
Several efforts have been conducted to alleviate these problems by enriching the distribution of the training domain.
% \begin{itemize}[leftmargin=*]
%     \item Distribution shifts on the graph can manifest simultaneously in both node attributes and topology, resulting in a substantial and intricate variation. 
%     \item Consequently, the variation poses challenges in capturing invariant information across different domains.   
% \end{itemize}
%\textcolor{red}{1-2 sentences here...}
% \textbf{State of the Art and its Limitations:}
% To tackle the aforementioned challenges, various initiatives have been undertaken to enhance the generalization of graph data in OOD settings by augmenting the quantity and diversity of training samples. 
These methods can be broadly categorized into three types: structure-wise, feature-wise, and mixed-type methods.
Structure-wise methods~\cite{park2021metropolis,wu2022knowledge} primarily focus on generating a greater variety of training topologies to cover some potential unobserved testing topologies.
Feature-wise methods~\cite{feng2020graph,kong2022robust} are devoted to generating a more diverse distribution of training attribute features.
These two types of methods specifically characterize distribution shifts either in node attributes or graph topologies.
Moreover, while mixed-type methods~\cite{you2020graph,sui2022adversarial} are introduced, these approaches primarily concentrate on graph-level tasks rather than node-level classification.
Node-level tasks are essential for capturing the local structures, features, and connections of individual nodes, often relying on information from nearby nodes or edges.
Nevertheless, graph-level tasks fail to take these factors into account, rendering them inappropriate for direct implementation in node-level tasks.

% \textbf{Our Approach and Contributions:} 
In response to the mentioned challenges and limitations, in this paper, we propose a principled framework, namely Graph Learning Invariant Domain genERation (\sysname{}), to address the OOD generalization problem for node-level classification on graphs.
Graph augmentation in \sysname{} contains two steps: (1) $G\rightarrow G'$: the distribution shift only occurs on the attribute matrix, but the topology structure remains, (2) $G'\rightarrow G''$: the distribution shift occurs on topology solely. 
%Furthermore, a domain-invariant classifier is learned using $G''$.
Specifically, in the initial step, the attribute of each node is disentangled into two factors in latent spaces: a domain-invariant semantic factor and a domain-specific variation factor.
By recombining semantic factors with randomly sampled variation factors for all nodes in a graph $G=(A,X)$, a novel graph $G' = (A, X')$ is obtained, where the node attributes $X'$ are generated from $X$ in $G$ but the adjacency matrix remains unchanged.
In the second step, we augment the topology through adversarial training.
Subsequently, a graph with distribution shift on both the node attribute matrix and topology, denoted as $G'' = (A', X')$, is generated.
Finally, \sysname{} learns the domain-invariant classifier for node label prediction by minimizing the empirical risk from multiple domains.
Our contributions are summarized as follows:
\begin{itemize}[leftmargin=*]
    % \item To the best of our knowledge, this is the first domain generalization framework proposed for node-level tasks on graphs that simultaneously considers distribution shifts in attribute matrix and topological structure. This approach aligns more closely with real-world graph data, providing a more comprehensive and realistic solution.
    \item We introduce a domain generalization framework for node-level tasks on graphs that simultaneously addresses distribution shifts in both attributes and topology. This approach aligns more closely with real-world graph data, offering a more comprehensive and realistic solution.

    \item We diversify the variation by modeling unseen variations in node attribute distribution and graphic topology. Furthermore, we minimize the semantic gap for each class by reducing the discrepancy of the variation in the representation space to learn the invariant information of nodes across domains.
    
    \item We conduct a set of comprehensive experiments on diverse real-world datasets for the node-level prediction task. Our results demonstrate the effectiveness of \sysname{} compared to state-of-the-art baselines.
    
\end{itemize}

\section{Related Works}
% \subsection{Out-of-distribution Generalization}
\textbf{Out-of-distribution (OOD) Generalization} aims to find a trained predictor using data drawn from a family of related training domains and then evaluated on a distinct and unseen test domain~\cite{robey2021model}.
Methodologically, OOD generalization algorithms can be broadly categorized into disentangled representation~\cite{kim2018disentangling,dittadi2020transfer}, causal representation~\cite{yang2021causalvae,sui2022causal}, domain generation~\cite{garg2021learn,zhou2022domain}, the combination of causal learning and invariant learning~\cite{gamella2020active,krueger2021out,xia2024learning} and stable learning~\cite{zhang2021deep,liu2021stable}.
Disentangled representation aims to learn representations that effectively separate relevant information factors from irrelevant ones.
Domain generalization aims to learn models with strong generalization performance on unpredictable target domains by leveraging multiple source domains with diverse distributions.
Following similar spirits, we minimize prediction variance across domains to capture invariant features but do not assume that the source domain comes from multiple environments.
% Therefore, it also applies to the scenario with a single graph input.

% \subsection{Out-of-distribution Generalization on Graphs}
\textbf{OOD Generalization on Graphs.} Despite extensive studies documenting success in addressing the OOD problem for Euclidean data, such as images or molecules, there has been relatively little attention given to graph-structured data, which is prevalent in real-world scenarios.
Recently, several graph enhancement methods for OOD generation on graphs have been introduced, which can be broadly categorized into three types: structure-wise augmentations~\cite{park2021metropolis,wu2022handling}, feature-wise augmentations~\cite{feng2020graph,kong2022robust}, and mixed-type augmentations~\cite{you2020graph,sui2022adversarial}.
Structure-wise augmentations aim at generating a greater variety of training topologies to cover some potential unobserved testing topologies.
Feature-wise augmentations are devoted to generating more diverse training attribute feature distribution.
The two methods described above solely consider a single type of distribution shift in isolation.
Indeed, in real-world scenarios, it is common for both attribute and topology shifts to occur simultaneously in graph-structured data.
Further, augmentation methods that consider both attributes and topology are proposed, but these methods are for graph-level tasks and are not suitable for node-level.
This limitation implies that there is scope for additional research or the creation of supplementary techniques to handle node-level issues on graph data.

\section{Preliminaries}
% PRELIMINARIES}

All notations used in this paper are listed in Table~\ref{tab.notations} of Appendix~\ref{notaions}. 

\subsection{Problem Setting and Formulation}
\label{section:Formulation}

% Given a network ${G} = ({A},{X})$, ${A} \in \mathbb{R}^{|\mathcal{V}| \times |\mathcal{V}|}$ and ${X}$ \textcolor{red}{$\in \mathbb{R}??$} denote the adjacency matrix and the node attribute feature matrix, respectively, where $|\mathcal{V}|$ is the number of nodes. 
% $\mathbf{x}_{v} \in \mathbb{R}^{m}$ denotes the feature vector of node ${v}$.
% \textcolor{red}{$\mathbf{y} = \{ y_v | v \in \mathcal{V}\}$ is a vector and $\mathbf{y}_v $ is the label of each node.}
%Let $\mathcal{G}$ be an input graph space and $\mathcal{Y}$ is the corresponding label space \textcolor{red}{for each node?}.
Denote a graph $\mathcal{G} = \{ G^{e} \}_{e=1}^{|\mathcal{E}|}$, where $G^{e} = (A^{e}, X^{e})$ is a subgraph with labels $\mathcal{Y}^e=\{y_v^e\}$ collected from the domain $e \in \mathcal{E}$ and $\mathcal{E}$ denotes the set of domains.
It is worth noting that we abuse this notation and all domains share the same label space.
${A}^e \in \mathbb{R}^{|\mathcal{V}^e| \times |\mathcal{V}^e|}$ denotes the adjacency matrix and ${X}^e \in \mathbb{R}^{|\mathcal{V}^e| \times d}$ denotes the node attribute matrix, where $|\mathcal{V}^e|$ is the number of nodes and $d$ is the dimension of node feature.
% In the setting of node-level domain generalization, a graph $G$ consists of a set of subgraphs 
% $\{ G^{e} \}_{e=1}^{|\mathcal{E}|}$, where $\mathcal{E}$ denotes the set of domains.
% A subgraph $G^{e} = (A^{e}, X^{e})$ is collected from the domain $e \in \mathcal{E}$.
% We assume each subgraph $G^{e}$ corresponds to a particular domain $e$, and the distributions of node attributes and topology in subgraphs vary across domains.
The distributions of node attributes and topology in subgraphs vary across domains.

The goal of node-level OOD generation on graphs is to improve the model's generalization ability for node label prediction across domains.
% Nevertheless, nodes exhibit variations in attribute distribution and topological structure in different domains.
% Concurrently, there exist intricate interrelations between nodes, making it impractical to consider each node independently.
To mitigate the challenges outlined in Section \ref{sec:intro}, we introduce the ego-graphs for each node $v\in\mathcal{V}$, capturing the surrounding influences on the centered node.
We define the node's ego-graph as $G_v^e = (A_v^e, X_v^e)$, where $A_v^e$ denotes the adjacency matrix and $X_v^e$ denotes the node feature matrix of the ego-graph for node $v\in\mathcal{V}^e$ in domain $e\in\mathcal{E}$.
An ego-graph ${G}_{v}^e$ of a node includes the center node (ego) $v$ and its ${L}$-hop neighbors, where ${L}$ is an arbitrary integer.
The set of edges comprises all the edges present in the ego-graph of these nodes.
Naturally, a graph will consist of a series of instances $ \{ ({G}_{v}^e, y_v^e) \}_{v=1}^{|\mathcal{V}^e|}$, where $y_v^e$ denotes the class label of the node ${v}$.

\begin{problem}[Node-Level OOD Generalization on Graphs]
\label{prob:problem1}
    Let $\mathcal{E}_{train}\subsetneq\mathcal{E}$ be a finite subset of training domains, and assume that for each $e\in\mathcal{E}_{train}$, we have access to a subgraph $G^e=(A^e,X^e)$, %A^e\in\mathbb{R}^{|\mathcal{V}^e|\times|\mathcal{V}^e|}$ 
    and the corresponding $ \{({G}_{v}^e, y_v^e) \}_{v=1}^{|\mathcal{V}^e|}$ sampled \textit{i.i.d} from $\mathbb{P}(G^e,{Y}^e)$, where ${G}_{v}^e=(A^e_v,X^e_v)$. Given a function class $\mathcal{F}$, a representation function $g:\mathcal{G}\rightarrow\mathbb{R}^n$ (\textit{e.g.,} GNN), and a loss function $\ell:\mathcal{Y}\times\mathcal{Y}\rightarrow\mathbb{R}$, the goal is to learn a predictor $f_c\in\mathcal{F}$ that minimizes the worst-case risk over $\mathcal{E}$.
    \begin{align}
    % \footnotesize
        \min_{f_c\in\mathcal{F}} \max_{e \in \mathcal{E}} \: \mathbb{E}_{(G^e_v,y^e_v)\sim\mathbb{P}({G}^e,{Y}^e)} \ell(f_c(g(G^e_v)),y^e_v)
        \label{eq.problem}
    \end{align}
\end{problem}

We assume $\mathcal{E}_{train}\subsetneq\mathcal{E}$ is accessible during training and $\mathcal{E}_{test}=\mathcal{E}\backslash\mathcal{E}_{train}$ is an inaccessible and unknown test domain.
% A more intuitive illustration is shown in Fig.~\ref{fig:ego-domain}.

\subsection{Assumptions}
% \begin{figure}[!t]
%   \centering
%   \includegraphics[scale=0.22]{figures/ego-domain.pdf}
%   \caption{Illustration of the problem setting. $G^{e_1}$, $G^{e_2}$ and $G^{e_3}$ are three subgraphs of $G$, sampled from three different domains ($e_1$, $e_2$ and $e_3$). 
%   %Distributions of node attributes and topology in subgraphs vary across domains. 
%   Taking the red node $v$ in $G^{e_1}$ as an example, its ego-graph $G_v^{e_1}$ with 2-hop neighbors ($L=2$) is shown on the left. }
%   \label{fig:ego-domain}
%   %\vspace{-5mm}
% \end{figure}
To tackle the problem mentioned above, we begin by establishing the following assumptions.

\begin{assumption}[Multiple Latent Factors]
\label{assumption_1}
Given a subgraph $G^{e}$ and the corresponding $\{(G_v^{e}, y_v^e)\}_{v=1}^{|\mathcal{V}^e|}$ from any domain $\forall e \in \mathcal{E}$.
Given a representation function (\textit{e.g.,} GNN) $g:\mathcal{G}\rightarrow\mathbb{R}^n$, we assume that the representation of each node $\mathbf{z}_v^e = g(G_v^{e})$ in $G^{e}$ is generated from 
\begin{itemize}
    \item a latent semantic factor $\mathbf{c} \in \mathcal{C}$, where $\mathcal{C} = \{\mathbf{c}_{y_{v}^{e}=0, } \mathbf{c}_{y_{v}^{e}=1}, \cdots,  \mathbf{c}_{y_{v}^{e}=C}\}$ denotes a semantic space and $C$ is the number of label categories;
    \item a latent variation factor $\mathbf{r}^{e}$, where $\mathbf{r}^{e} = \{ \mathbf{r}_a^{e}, \mathbf{r}_t^{e} \}$  is specific to the individual domain $e$.
    Herein, $\mathbf{r}_a^{e}$ is the variation factor on attribute and $\mathbf{r}_t^{e}$ is the variation factor on topology.
    The variation factor $\mathbf{r}^{e} $ controls the change of domain $e$.
    % In other words, all nodes in the same domain have the same variation factor.    
    
\end{itemize}
The above assumption indicates that for any two node representations, $\bar{v}_1:=\{{\mathbf{z}}_{v_1}^{e}, y_{v_1}^{e}\}$ sampled from $G^{e}$ and $\bar{v}_2:=\{{\mathbf{z}}_{v_2}^{e'}, y_{v_2}^{e'}\}$ sampled from $G^{e'}$
\begin{itemize}
    \item if $e=e'$ and $y_{v_1}^{e} = y_{v_2}^{e'}$, then $\bar{v}_1$ and $\bar{v}_2$ share the same $\mathbf{c}$ and $\mathbf{r}^e$;
    \item if $e =e'$ and $y_{v_1}^{e} \neq y_{v_2}^{e'}$, then $\bar{v}_1$ and $\bar{v}_2$ share the same $\mathbf{r}^e$, but different $\mathbf{c}$;
    \item if $e \neq e'$ and $y_{v_1}^{e} = y_{v_2}^{e'}$, then $\bar{v}_1$ and $\bar{v}_2$ share the same $\mathbf{c}$, but different $\mathbf{r}^e$ and $\mathbf{r}^{e'}$;
    \item if $e \neq e'$ and $y_{v_1}^{e} \neq y_{v_2}^{e'}$, then neither $\mathbf{c}$ or $\mathbf{r}^e$ and $\mathbf{r}^{e'}$ are the same for $\bar{v}_1$ and $\bar{v}_2$.   
\end{itemize}
\end{assumption}

Note that Assumption~\ref{assumption_1} is similar to \cite{liu2017unsupervised,huang2018multimodal} and has been extensively employed in interpretability research~\cite{liu2021learning,zhang2022towards}.
We extend the assumptions of existing works by introducing two latent factors to focus on graph data.

% \textcolor{red}{add 1-2 sentences here related to Assumption 1...}

\begin{assumption}[Invariance on Graphs]
\label{assumption_2}
We assume that inter-domain variation is solely characterized by domain shift in the marginal distributions $\mathbb{P}({{Z}}^e), \forall e\in\mathcal{E}$.
We assume that $\mathbb{P}({Y}^e | {{Z}}^e) $ is stable across domains.
Given a domain transformation function $\mathcal{T}$, for any ${\mathbf{z}} \in \mathbb{R}^m$, $y \in \mathcal{Y}$, it holds that:
\begin{equation}
    % \footnotesize
    \begin{aligned}
          \mathbb{P}({Y}^{e} = y | {{Z}}^{e} = {\mathbf{z}}^{e}) = \mathbb{P}({Y}^{e'}=y | {{Z}}^{e'} = \mathcal{T}({\mathbf{z}}^{e}, e') ), 
          \forall e,e' \in \mathcal{E}, e \neq e'
        \end{aligned}
        \label{eq.first}
\end{equation}
\end{assumption}

\begin{figure}[t!]
  \centering
  \includegraphics[width=0.8\linewidth]{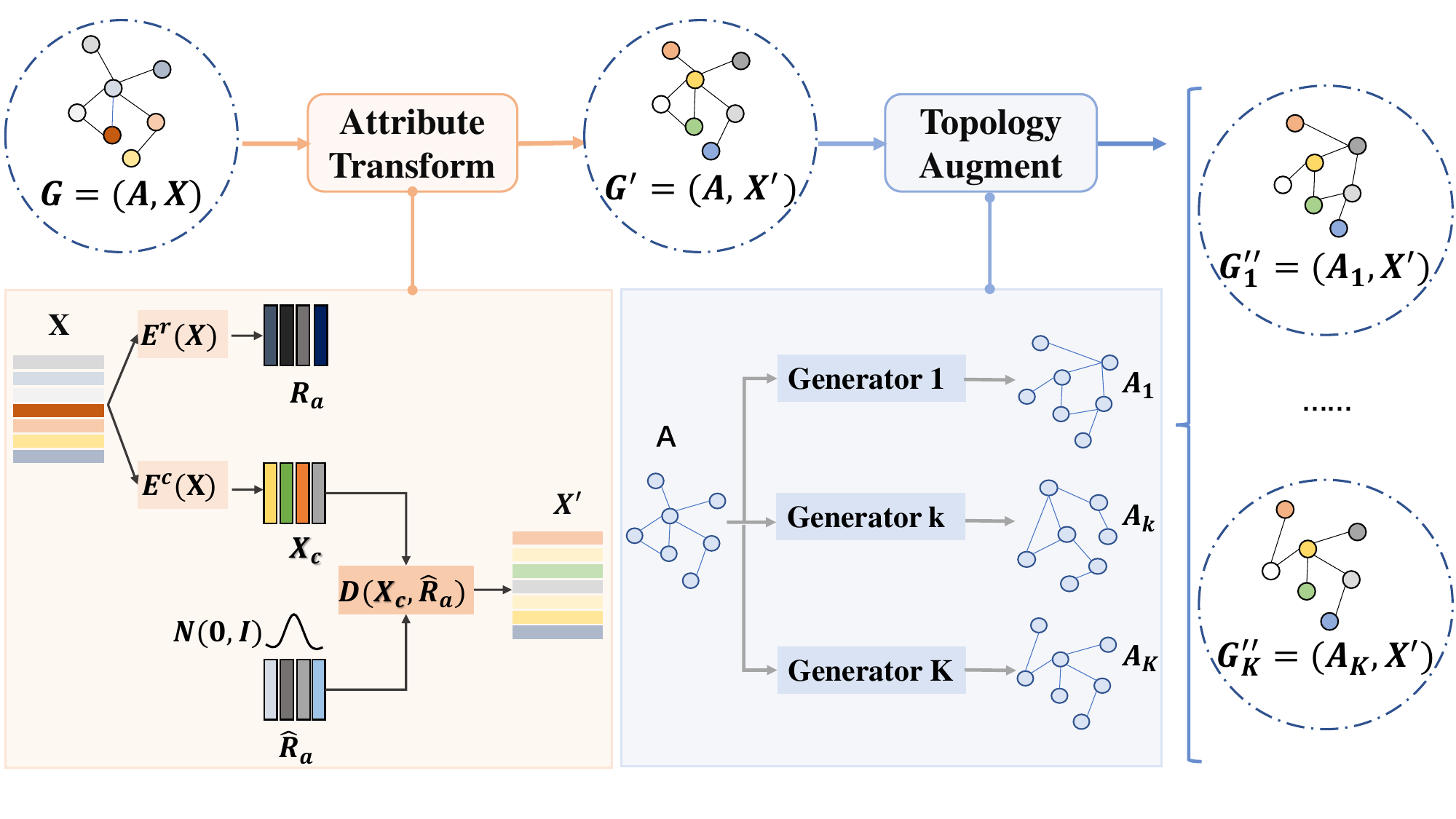}
  % \vspace{-8mm}
  \caption{An overview of the \sysname{} framework. 
  First, \sysname{} generates a $G'$ that has the same topology, but a different attribute distribution with the input graph $G$.
  In this step, it learns a semantic feature encoder $E^c$, a variation feature encoder $E^r$, and a decoder $D$.
  Then it generates node attributes $X'$ in new synthetic domains, wherein only the variation factor is replaced with a sampled one from $\mathcal{N}(0,\mathbf{I})$. 
  Second, it learns generators that can generate $K$ adjacent matrix with structure as differently as possible. 
  Then we can generate $K$ graphs $\{G_k^{''}=(A_k, X')\}_{k=1}^K$ with different $X$ and $A$ compared with $G$.
  % Note, that the color represents attribute characteristics, and the shape of the connection between nodes represents the topology structure.
  }
  \label{fig:framework}
  % \vspace{-3mm}
\end{figure}

Some recent works~\cite{wu2022knowledge,kong2022robust} for node-level OOD generation on graphs problem have considered either topology variation~\cite{park2021metropolis,wu2022knowledge} or node attribute variation~\cite{feng2020graph,kong2022robust} only.
They do not take into account the variations that occur simultaneously.
Based on Assumption~\ref{assumption_1}, distribution shift across domains is determined by $\mathbf{r}^e, \forall e\in\mathcal{E}$ that includes variation factors for both node attribute $\mathbf{r}_a^e$ and topology $\mathbf{r}_t^e$.
Expanding on the insights from the existing literature~\cite{zhang2022towards}, we emphasize that inter-domain variation is solely characterized by the domain shift on $\mathbb{P}({{{Z}}^e})$ due to $\mathcal{T}$.

\section{Methodology}

In this section, we will present \sysname{} in detail.
Fig.~\ref{fig:framework} shows an overview of \sysname{} and its learning process.
The algorithmic pseudocode of the whole framework is given at Algorithm~\ref{whole_algorithm}.

% \sysname{} is composed of two parts. 
% Firstly, it is devoted to generating a new $G'$ that retains the same topology as $G$ but with a different attribute distribution.
% % \sysname{} learns two encoders and a decoder to create a variable attribute feature matrix called $X'$. 
% % From this matrix, it then generates a new graph $G' = (A, X')$.
% %$X' = [\mathbf{x'_1}, \mathbf{x'_2}, \cdots, \mathbf{x'_{|V|}}] $ 
% %$G' = (A, X')$ 
% Secondly, it trains multiple graph generators to produce $K$ distinct topology matrices, denoted as $A_k$ ($k=1, \cdots, K$).
% % By integrating the generated attribute feature matrix and the topology matrices, a variety of graphs $ \{ G_1^{''}, \cdots, G_K^{''} \}$ can be  generated, %\textit{i.e.,} $G{''}=\{ G''_1. \cdots, G''_K\}$ 
% % where ${G}_k^{''}=({A}_k,{X^{'}})$.
% % $G$ and each $G''$ have different distributions of node attributes and topology.
% % We visualize our framework in Fig.~\ref{fig:framework}.
% Finally, \sysname{} would like to learn a domain-invariant predictor $f_c \in \mathcal{F}$ using the generated graphs.
% This will be achieved by minimizing the worst-case risk across all generated graphs. 

\subsection{Attribute Matrix Transformation}
\label{section_Attribute}

% Herein, we demonstrate the generation of the graph $G'=(A, X')$ from $G=(A,X)$ where we only focus on the distribution shift on node attributes and keep the adjacency matrix $A$ unchanged.

According to Assumption~\ref{assumption_1},  an attribute vector $\mathbf{x}^e$ of a node of a specific domain $e$ is generated from a latent semantic factor $\mathbf{x}_{c}$, and a variation factor $\mathbf{r}_a^{e}$ that is specific to the domain with $A$ unchanged.
% As stated in Section~\ref{section:Formulation}, a given graph $\mathcal{G}$ contains multiple domains where each corresponds to a variation factor. 
We denote $R_a =[\mathbf{r}_a^{e_1}, \cdots, \mathbf{r}_a^{e_{|\mathcal{E}|}}]$ as the variation matrix for transformation from $G$ to $G'$.

We encode the attribute feature vector $\mathbf{x}$ of each node in $G$ into a latent semantic factor $\mathbf{x}_c \in \mathbb{R}^{d_c} $ and a latent variation factor $\mathbf{r}_{a}^e \in \mathbb{R}^{d_r}$.
Our objective is to learn a domain-invariant semantic feature encoder ${E}^{c}$, domain-specific variation feature encoder ${E}^{r}$, and a decoder $D$ to generate various attribute feature vectors for each node.
To achieve this, we generate node attributes in new synthetic domains, wherein only the variation factor is replaced with a randomly sampled one.
We draw some latent variation factor $\hat{R}_a$ from the prior distribution $q(\hat{{R}}_a) \sim \mathcal{N}(0,\mathbf{I})$.
These generated $\hat{\mathbf{r}}_a$ and $\mathbf{x}_c$ are used to generate multiple attribute feature vectors preserving invariant information, \textit{i.e.,} $\mathbf{x'} = D(\mathbf{x}_c, \hat{\mathbf{r}}_a)$.
In the same way, we also encode the generated $\mathbf{x'}$ into a latent semantic factor $\mathbf{x'}_c$ and a latent variation factor $\mathbf{r}_{a}^{'e}$.
Herein, $\mathbf{x}_c$ and $\mathbf{x'}_c$ should theoretically be equivalent.
A new graph, $G'=(A, X')$, with the same topology as $G$ but different attribute distribution of node, can be generated, where $X' = [\mathbf{x'}_1, \mathbf{x'}_2, \cdots, \mathbf{x'}_{|\mathcal{V}|}]$.
To accomplish the above objectives, we define the loss function of this process.
The loss comprises a bidirectional reconstruction loss to guarantee that the encoders and decoder work inversely, and an adversarial loss to align the distribution of the translated and original feature matrix.

\textbf{Bidirectional reconstruction loss.} 
To learn mutually inverse pairs of encoder and decoder, we utilize objective functions that encourage reconstruction in both the matrix $\to$ latent $\to$ matrix and latent $\to$ matrix $\to$ latent directions.
\begin{itemize}[leftmargin=*]
    \item \textit{Matrix reconstruction.} We use encoding and decoding to reconstruct the attribute feature matrix that was sampled from the data distribution.
    The objective function can be formulated as the following:
    \begin{equation}
    % \footnotesize
        \begin{split}
          \mathcal{L}_{rec}^{x} = \mathbb{E}_{X \sim \mathbb{P}({X})} \lbrack \Vert {D} ({X_c}, R_a), X \Vert_1 \rbrack,
        \end{split}
        \label{eq.Recon_x}
    \end{equation}
    where $X_c = E^c(X)$ and $R_a = E^r(X)$.
    \item \textit{Latent reconstruction.} Given a latent semantic factor and a variation factor sampled from the prior distribution $\mathcal{N}(0,\mathbf{I})$, we should attempt to reconstruct it after decoding and encoding.
     \begin{equation}
     % \footnotesize
        \begin{split}
          \mathcal{L}_{rec}^{c}=\mathbb{E}_{X_c \sim \mathbb{P}({C}),\hat{R}_a \sim q(\hat{{R}}_a) } \lbrack \Vert E^{c}({D} ({X_c}, \hat{R}_a)) - {X_c} \Vert_1 \rbrack,
        \end{split}
        \label{eq.Recon_C}      
    \end{equation}
      \begin{equation}
       % \footnotesize
        \begin{split}
          \mathcal{L}_{rec}^{r}=\mathbb{E}_{X_c \sim \mathbb{P}({C}),\hat{R}_a \sim q(\hat{{R}}_a) } \lbrack \Vert E^{r}({D} ({X_c}, \hat{R}_a)) - \hat{R}_a \Vert_1 \rbrack,
        \end{split}
        \label{eq.Recon_S}
    \end{equation} 
where $q(\hat{{R}}_a)$ is the prior $\mathcal{N}(0,\mathbf{I})$, $ \mathbb{P}({C})$ is given by $X_c=E^c(X)$.
The semantic feature reconstruction loss $\mathcal{L}_{rec}^{c}$ has the effect of encouraging the translated matrix to preserve the semantic feature of the input matrix.
The variation feature reconstruction loss $\mathcal{L}_{rec}^{r}$ is capable of generating diverse outputs given diverse variation factors.
\end{itemize}

\textbf{Adversarial loss.}
To guarantee the matrix generated by our model is indistinguishable from the real feature matrix in the target domain, we employ \textit{maximizing mutual information} to match the distribution of the translated matrix to the real data.
We employ a discriminator $\Psi$: $\mathcal{G} \rightarrow \mathbb{R}$, where $\Psi(D(X_c,\hat{R}_a))$ represents the probability scores of similarity between translated and real matrix.
Negative samples $X'$ for $\Psi$ are generated by pairing the semantic latent factor with a latent variation factor sampled from the $\mathcal{N}(0,\mathbf{I})$.
The objective is formulated as follows:
\begin{equation}
% \footnotesize
        \begin{split}
          \mathcal{L}_{Di}=\mathbb{E}_{X'\sim \mathbb{P}({X'})}  \lbrack \log (1-\Psi(X'))\rbrack + \mathbb{E}_{X\sim \mathbb{P}({X})} \lbrack \log \Psi(X) \rbrack,
        \end{split}
        \label{eq.GAN}
\end{equation}
where $X^{'} = D(X_c,\hat{R}_a)$.
This approach effectively maximizes the mutual information between $X'$ and $X$.

\textbf{Total loss.} 
We jointly train the encoders, decoder, and discriminators to optimize the final objective, which is a weighted sum of the adversarial loss and the bidirectional reconstruction loss terms.
\begin{equation}
    % \small
    % \footnotesize
    \begin{aligned} 
     \mathcal{L} =  \mathcal{L}_{Di} 
    + \lambda_{x} \mathcal{L}_{rec}^{x} 
    +  \lambda_{c} \mathcal{L}_{rec}^{c} 
    + \lambda_{s} \mathcal{L}_{rec}^{r}
    \label{eq.totalLoss},
\end{aligned}
\end{equation}
where $\lambda_x$, $\lambda_c$, $\lambda_s$ are the weight coefficient of each part.

\subsection{Topology Augmentation}
Given a graph $G'=(A,X')$ generated from the original graphs $G$, we now demonstrate the generation of a sequence of graph $ \{G_k^{''}=(A_k,X')\}_{k=1}^K$. 
In contrast to $G'$, each $G_k^{''}$ in the sequence maintains the node attribute $X'$ and changes the topological structure from $A$ to $A_k$.
%with distribution shift on both attribute feature and topology structure compared with the input graph $G$.
% We take the generated $G'$ as input for this process.
We define $K$ graph generators $\{\hat{g}_{w_k}\}_{k=1}^K$.
%$K$ is the number of generated topology matrix $A$.
Herein, the goal of our framework is to learn generators  $\{\hat{g}_{w_k}\}_{k=1}^K$ with parameters $\{w_1, \cdots, w_K\}$, where $\hat{g}_{w_k}(G') = G''_k = (A_k, X')$.
These generators are devoted to generating $K$ graphs with the least possible similarity in their node topologies.
For simplicity, we use $G^k$ to represent the $G_k^{''}$ in this section.

To begin, we define the mean value of prediction losses for all nodes in a generated graph as $l_k = \frac{1}{|\mathcal{V}|} \sum_{v\in\mathcal{V}} \ell(f_c(g(G_v^k), y_v^k))$, where $G^k = \hat{g}_{w_k}(G')$ and $G_v^k$ is corresponding ego-graph of node $v$ in $G^k$.
To enhance topological differences among graphs, we train generators by maximizing the variance of $l_k$ across them:
\begin{equation}
% \footnotesize
    \begin{split}
     \max_{w_1, \ldots,w_K} \mathbb{V}({l_k): (1\le k \le K}).
    \end{split}
    \label{eq.gwk}
\end{equation}
%where $L(g_{w_k}(G'), \mathbf{y}) = L(G^{''}, \mathbf{y})$ is the prediction loss with input graph $G'$.

% In this section, we will introduce how to optimize the graph generator, i.e., $g_{w_k}$ in Equation~\ref{eq.Loss} in detail.
% The optimization process is following \cite{wu2022handling}.
Inspired by \cite{jin2020graph}, we use the data augmentation method of adding and subtracting edges to learn the generator $\hat{g}_{w_k}$.
We define a Boolean matrix $\bar{R}_t^k = \{0,1\} ^{|\mathcal{V}| \times |\mathcal{V}|}$ ($k=1, \dots, K$) and the supplement matrix of $A$ as $\bar{A} = \mathbf{1}\mathbf{1}^T - I - A$, where $I$ is an identity matrix and $\mathbf{1}$ is a vector of length $|\mathcal{V}|$ with each element is $1$.
Element in the matrix $\bar{R}_t$ represents whether the two nodes in the corresponding position are connected by an edge.
Then a new topology matrix is obtained by $A_k = A + \bar{R}_t^k \circ (\bar{A} -A)$, where $\circ$ denotes the element-wise product.
However, $\bar{R}_t^k$ is indifferentiable and the objective function is non-convex.
To solve this dilemma, we introduce the policy gradient method to optimize the $\hat{g}_{w_k}$, which regards the generation of graphs as a decision process and edge operation as actions.
We define a parameterized matrix set $\mathcal{B} = \{B_1, B_2, \dots, B_K \}$, where $B_k = \{\beta_{ij}^k\}$.
Take the $i$-th node as an example, the probability of editing the edge between node $i$ and node $j$ for the $i$-th node in view $k$ is $p(\sigma_{ij}^k) = \frac{exp(\beta_{ij}^k)}{\sum_{j'\in \mathcal{V}} exp(\beta_{ij'}^k)}$.
Then we execute $s$ actions $\{ s_{ij_{t}}^k\}_{t=1}^s$ from a multinomial distribution $\Omega(p(\sigma_{i1}^k), p(\sigma_{i2}^k), \cdots, p(\sigma_{i|\mathcal{V}|}^k), \cdots, p(\sigma_{|\mathcal{V}||\mathcal{V}|}^k))$, which determines the non-zero element in row $i$ of $\bar{R}_t^k$.
Moreover, we define the reward function $Q$ as the inverse of the loss in Eq.~\eqref{eq.gwk}.
Herein, we utilize REINFORCE~\cite{zhang2021sample} algorithm to optimize the $w_k$ with the gradient $\bigtriangledown_{w_k}\log p_{w_k}(A_k)Q$, where $w_k = B_k$ and $p_{w_k}(A_k) = \prod_{i=1}^{|\mathcal{V}|} \prod_{t=1}^s p(s_{ij_{t}}^k)$.

\subsection{Invariant Representation Learning}
Once we complete the initial two steps, we produce $K$ graphs $G''$ that display a different distribution of attributes and topology of nodes.
Then we will extract the ego-graph $G_v^{''}$ for each node $v$ in $G''$ and generate a series of instances $ \{ ({G}_{v}^{''}, y_v^{''}) \}_{{v} \in {V}^{''}}$.
%Under the circumstances, invariant feature information cannot be intuitively defined for nodes, because of (1) the ego-graph consists of $L$-order neighbors of the center node and relations between them, which introduce complex hierarchical structure and (2) the nodes' quantity of each order is inconsistent.
%To mitigate the issue, we extract the ego-graph for each node.
%We define a predictor $f_c \in \mathcal{F}$, a representation function $g$ (\textit{e.g.,} GNN), and a loss function $\ell:\mathcal{Y}\times\mathcal{Y}\rightarrow\mathbb{R}$.
For Problem~\ref{prob:problem1}, we utilize GNN as the representation function $g$, which learns the representation of each node by iteratively aggregating the neighbors' information of the center node in the ego-graph.
Please note that in this context, $G''$ refers to a graph that has been sampled from various domains. 
To maintain consistency, we will continue to use $e$ to represent a distinct domain in the following text.

%Inspired by~\cite{leman1968reduction} and based on Assumption~\ref{assumption_2}, we define $h^*$ as an encoder: $\mathbf{r}_v=h^*(G_v)$, a classifier $f_c$: $\hat{y}=f_c(\mathbf{r}_v)$ and $l (\cdot , \cdot)$ denotes a loss function.
%Denote $e \in \mathcal{E}$ as the environment variable.
%The node-level OOD problem on graphs can be formalized:

% \begin{equation}
%         \begin{split}
%          \min_{\theta} \max_{e \in \varepsilon} \mathbb{E}_{G \sim p(G|e=e)} [\frac{1}{|V|} \sum_{v \in V} \mathbb{E}[l(f_c(h_v),y_v);\theta]].
%         \end{split}
%         \label{eq.invarant_2}
% \end{equation}

We assume that when the variance of empirical risk across different environments is smaller, the model tends to make predictions with invariant characteristics, \textit{i.e.,} when the objective $\min \mathbb{V}_e[R(e)]$ reaches the optimum.
So we get the following objective:
\begin{equation}
% \footnotesize
        \begin{split}
         \min &\mathbb{V}_e[\ell ( f_c(g(G_v^e)),y_v^e)] + \alpha \mathbb{E}_e[\ell(f_c(g(G_v^e)),y_v^e)] \\
         %\text{s.t. } &\max_{w_1, \ldots,w_K} \mathbb{V}({l_k): (1\le k \le K}),  
        \end{split}
        \label{eq.invarant_3}
\end{equation}

% \begin{equation}
% \footnotesize
%         \begin{split}
%          \min_{\boldsymbol{\theta}} \: &\mathbb{V}_e[\ell ( f_c(g(G_v^e; \boldsymbol{\theta})),y_v^e)] + \alpha \mathbb{E}_e[\ell(f_c(g(G_v^e; \boldsymbol{\theta})),y_v^e)] \\
%          %\text{s.t. } &\max_{w_1, \ldots,w_K} \mathbb{V}({l_k): (1\le k \le K}),  
%         \end{split}
%         \label{eq.invarant_3}
% \end{equation}
%where $ L(G^e,Y^e) = \frac{1}{|V_e|} \sum_{v \in V} l(f_c(G_v),\mathbf{y}_v^e)$.
where $\alpha$ is a weight coefficient.
%where $L(g_{w_k}(G^e),Y^e)=L(G^{''},Y^{''})=\frac{1}{|V|} \sum_{v \in V}l(f_c(h^*(G_v{''})),y_v^{''})$.
%where $L(g_{w_k}(G^e),Y^e)=\frac{1}{|V|} \sum_{v \in V}l(f_c(h^*(G_v{''})),y_v^{''})$.
\begin{algorithm}[t!]
    \caption{\textbf{\sysname{}.}} \label{whole_algorithm}
    \begin{algorithmic}[1]
    \STATE \textbf{Input:} an input graph $G=(A,X)$ and labels $\mathcal{Y}$, initialized parameters $\boldsymbol{\theta}$ of GNN, learning rates $l_f$, $l_g$, hyper-parameter $\alpha$, graph editers $\{w_k\}_{k=1}^K$.
    \STATE \textbf{Repeat}: 
    \STATE \qquad $X' \leftarrow$  \textsc{Generate}($X$)
    \STATE \qquad $\{G_k^{''} \}_{k=1}^K\leftarrow $ \textsc{Augmentation}(A,X')
    \STATE \qquad $Loss~\mathcal{L} \leftarrow $ compute using Eq.~\eqref{eq.invarant_3}
    \STATE \qquad Update $\boldsymbol{\theta} \leftarrow \boldsymbol{\theta}-l_f\bigtriangledown_{\boldsymbol{\theta}} \mathcal{L}(\boldsymbol{\theta})$
    \STATE \textbf{Until} convergence
    \STATE
    \STATE \textbf{Procedure} \textsc{Generate}($X$)
    \STATE \qquad \textbf{Repeat}: 
    \STATE \qquad \qquad $(X_c,X_{r_a}) \leftarrow E^c(X),E^r(X)$
    \STATE \qquad \qquad Sample $\hat{\mathbf{r}}_a \sim \mathcal{N}(0,\mathbf{I})$
    \STATE \qquad \qquad $X^{'} = D(X_c,\hat{\mathbf{r}}_a)$
    \STATE \qquad \qquad Compute loss $\mathcal{L}_{rec}^{x}$, $\mathcal{L}_{rec}^{c}$, $\mathcal{L}_{rec}^{r}$, $\mathcal{L}_{Di}$, and $\mathcal{L}$ in  \\ \ \  \  \ \ \  \ \ \  \ \ \  \ \ \ \ Eq.~\eqref{eq.Recon_x}, Eq.~\eqref{eq.Recon_C}, Eq.~\eqref{eq.Recon_S}, Eq.~\eqref{eq.GAN} and Eq.~(\ref{eq.totalLoss})
    % \STATE \qquad \qquad Compute $\mathcal{L}_{recon}^{c}$ in \eqref{eq.Recon_C}
    % \STATE \qquad \qquad Compute $\mathcal{L}_{recon}^{r}$ in \eqref{eq.Recon_S}
    % \STATE \qquad \qquad Compute $\mathcal{L}_{Dis}$ in ~\eqref{eq.GAN}
    % \STATE \qquad \qquad Compute $\mathcal{L}$ in (\ref{eq.totalLoss})
    \STATE \qquad \textbf{Until} convergence
    \STATE \textbf{End procedure}
    \STATE
    \STATE \textbf{Procedure} \textsc{Augmentation}($G'$)
    \STATE \qquad \textbf{For} $t=1$, $\cdots$, $T$ \textbf{do}
    \STATE \qquad \qquad Generated graphs $G^k = (A_k, X')$ from graph \\ \ \  \  \ \ \ \ \ \ \ \ \ \ \ \ \ generator $\hat{g}_{w_k}, k=1, \cdots, K$
    %\STATE \qquad \qquad Compute loss $\mathcal{L}(w) = \mathbb{V}(l_k)$
    \STATE  \qquad \qquad Compute loss $\mathcal{L}(w) = \mathbb{V}(l_k)$ in Eq.~\eqref{eq.gwk}
    \STATE  \qquad \qquad Update $w_k \leftarrow w_k + l_g\bigtriangledown_{w_k}\log p_{w_k}(A_k) \mathcal{L}(w)$
    \STATE \qquad \textbf{End for}
    \STATE \textbf{End procedure}

    % \STATE \textbf{procedure} Generate($G^{''}$)
    % \STATE \qquad \textbf{while} not converged or maximum epochs not reached \textbf{do}
    % \STATE \qquad \textbf{for} $t = 1, \cdots, T$ \textbf{do}
    % \STATE \qquad \qquad Obtain modified graphs $G_k = (A_k, X')$ from graph editer $g_{w_k}$, $k = 1, \cdots, K$
    \end{algorithmic}
\end{algorithm}

\section{Theoretical Analysis}
%In this section, we present a theoretical analysis of Problem~\ref{prob:problem1}.
Under Assumption~\ref{assumption_1} and Assumption~\ref{assumption_2}, if we restrict the domain $\mathcal{F}$ to the set of domain-invariant predictors, then Problem~\ref{prob:problem1} is equivalent to the following constrained optimization problem:
\begin{equation}
% \footnotesize
        \begin{split}
        &P^{*} \triangleq 
        \min_{f_c \in \mathcal{F}} \: R(f_c) \triangleq \mathbb{E}_{\mathbb{P}({{Z}}, {Y})} \ell( f_c({\mathbf{z}}), y) \\
        &\text{s.t. } \: f_c({\mathbf{z}}) = f_c(\mathcal{T}({\mathbf{z}}, e) )
         \quad a.e. \: {\mathbf{z}} \sim \mathbb{P}({{Z}}), \forall e \in \mathcal{E}
        \end{split}
        \label{eq.optimazation_1}
\end{equation}
where $\mathbf{z}=g(G_v^e)$ and $R(f_c)$ is the statistical risk of a predictor $f_c$.
% Enforcing the equality constraint is one of the most fundamental challenges.
However, enforcing a hard invariance constraint on the class $\mathcal{F}$ is not clear a priori.
To render it more tractable, we introduce the following relaxation of Eq.~\eqref{eq.optimazation_1}:
\begin{equation}
    % \footnotesize
    \begin{aligned}
        &P^{*}(\gamma) \triangleq \min_{f_c \in \mathcal{F}} \: R(f_c) \\
        % \footnotesize
        &\text{s.t. } \: \mathcal{L}^e(f_c) \triangleq \mathbb{E}_{\mathbb{P}({{Z}})}h(f_c({\mathbf{z}}), f_c(\mathcal{T}({\mathbf{z}}, e)))\leqslant \gamma, \forall e \in \mathcal{E},
    \end{aligned}
    \label{eq.optimazation_2}
\end{equation}
where $\gamma > 0$ is a fixed margin that regulates the difference between classifiers in various environments and $h$: $\mathcal{P}(\mathcal{Y}) \times \mathcal{P}(\mathcal{Y}) \to \mathbb{R}_{\geqslant 0} $ is a distance metric between different distributions.
By relaxing the equality constraints in Eq.~\eqref{eq.optimazation_1} to the inequality constraints in Eq.~\eqref{eq.optimazation_2} and under appropriate conditions on $\ell$ and $h$, Eq.~\eqref{eq.optimazation_2} can be characterized using the constrained PAC learning framework.
This framework is capable of providing learnability guarantees on constrained statistical problems.
% Indeed, it is worth noting that the conditions we impose on $h$ are not restrictive, such as the KL-divergence, as well as the more general family of $f$-divergences. 
Furthermore, when assuming that the perturbation function $P^*(\gamma)$ is L-Lipschitz continuous and $\gamma$ is strictly greater than zero, we have that $|P^* - P^*(\gamma)| \leqslant L\gamma$, which means the discrepancy between the Eq.~\eqref{eq.optimazation_1} and Eq.~\eqref{eq.optimazation_2} is relatively small when a small value of $\gamma$ is selected.
Eq.~\eqref{eq.optimazation_2} poses an infinite-dimensional constrained optimization problem over a functional space $\mathcal{F}$, rendering it intractable.
Therefore, to address this challenge, we follow the standard convention by leveraging a finite-dimensional parameterization of $\mathcal{F}$.
We can quantify the approximation power of such a parameterization using the following definition:

\textbf{Definition 1} ($\epsilon$-parameterization). Let $\mathcal{H} \subseteq \mathbb{R}^p$ be a finite-dimensional parameter space.
For $\epsilon > 0$, a function $\varphi: \mathcal{H} \times \mathcal{Z} \rightarrow \mathcal{Y}$ is said to be an $\epsilon$-parameterization of $\mathcal{F}$ if it holds that for each $f_c \in \mathcal{F}$, there exits a parameter $\boldsymbol{\theta}\in \mathcal{H}$ such that $\mathbb{E}_{\mathbb{P}({{Z}})} \parallel \varphi(\boldsymbol{\theta}, {\mathbf{z}}) - f_c({\mathbf{z}}) \parallel_\infty \leqslant \epsilon$.
Then, optimizing Eq.~\eqref{eq.optimazation_1} becomes tractable when dealing with a finite parametric space $\mathcal{U}_\epsilon:=\{ \varphi(\boldsymbol{\theta}, \cdot): \boldsymbol{\theta} \in \mathcal{H}\} \subseteq \mathcal{F}$.
To solve the above parameterization problem, we consider the following saddle-point problem:
\begin{equation}
    % \footnotesize
    \begin{aligned}
        D^{*}_{\epsilon}(\gamma) \triangleq \max_{\tau \in \mathcal{P}(\mathcal{E})} \min_{\boldsymbol{\theta} \in \mathcal{H}} R(\boldsymbol{\boldsymbol{\theta}}) + \int_{\mathcal{E}} [\mathcal{L}^e(\boldsymbol{\theta}) - \gamma]d \tau(e),
    \end{aligned}
    \label{eq.optimazation_3}
\end{equation}
where $\mathcal{P}(\mathcal{E})$ denotes the space of normalized probability distributions over $\mathcal{E}$ and $\tau \in \mathcal{P}(\mathcal{E})$ denotes the (semi-infinite) dual variable.
Let $\gamma > 0$ and there exists a small universal constant $t$, the optimality gap between Eq.~\eqref{eq.optimazation_2} and Eq.~\eqref{eq.optimazation_3} can be explicitly bounded as follows:
\begin{equation}
% \footnotesize
        \begin{split}
    P^{*}(\gamma) \leqslant D^{*}_{\epsilon}(\gamma) \leqslant P^{*}(\gamma) + \epsilon t (1 + \parallel \tau^*_{pert} \parallel_{1}),
        \end{split}
        \label{eq.optimazation_4}
\end{equation}
where $\tau^*_{pert}$ is the optimal dual variable for a perturbed version of Eq.~\eqref{eq.optimazation_2} in which the constraints are tightened to maintain a margin $\gamma-t\epsilon$.
This reveals that the difference between $P^{*}(\gamma)$ and $D^{*}_{\epsilon}(\gamma)$ is small when employing the $\epsilon$-parameterization on $\mathcal{F}$.
Nevertheless, we do not have access to $ \mathcal{E} \backslash \mathcal{E}_{train}$.
In practice, it is ubiquitous to solve optimization problems such as Eq.~\eqref{eq.optimazation_3} over a finite sample of $N$ data points.
Thus, given $\{ ({\mathbf{z}}_{v_i}, y_{v_i}) \}_{i=1}^N$ drawn \textit{i.i.d} from $\mathbb{P}({{Z}}, {Y})$, we consider the empirical counterpart of Eq.~\eqref{eq.optimazation_3}:
\begin{equation}
    % \footnotesize
    \begin{aligned}
     D^{*}_{\epsilon, N, \mathcal{E}_{train}}(\gamma) & \triangleq \max_{\tau(e) \geq 0, e \in \mathcal{E}_{train}} \min_{\boldsymbol{\theta} \in \mathcal{H}} \Gamma(\boldsymbol{\theta}, \tau) \\
     & \triangleq \widehat{R}(\boldsymbol{\theta}) + \frac{1}{|\mathcal{E}_{train}|} \sum_{e \in \mathcal{E}_{train}}[\widehat{\mathcal{L}}^e(\boldsymbol{\theta}) - \gamma] \tau(e),
    \end{aligned}
    \label{eq.optimazation_5}
\end{equation}
where $\widehat{R}(\theta) := \frac{1}{N}\sum_{i=1}^N l(\varphi(\boldsymbol{\theta},{\mathbf{z}}_{v_i}), y_{v_i})$ and $\widehat{\mathcal{L}}^e(\boldsymbol{\theta}) := \frac{1}{N} \sum_{i=1}^N d(\varphi(\boldsymbol{\theta}, {\mathbf{z}}_{v_i}), \varphi(\boldsymbol{\theta}, {\mathbf{z}}_{v_i}^e))$.
$\Gamma(\boldsymbol{\theta}, \tau)$ is the empirical Lagrangian. 
We can explicitly bound the duality gap between the solution to Eq.~\eqref{eq.optimazation_5} and the original Problem \ref{prob:problem1} in the following way.

%\textbf{Theorem 1 } (Data-dependent duality gap).
\begin{theorem}[Data-dependent duality gap]
\label{theorem}
Given $\epsilon > 0$ and let $\varphi$ be an $\epsilon$-parameterization of $\mathcal{F}$.
Under mild regularity assumptions on $\ell$ and $h$ and assuming that $\mathcal{U}$ has finite VC-dimension, with probability $1-\rho$ over the $N$ samples representations from $\mathbb{P}({{Z}}, {Y})$, we have that
\begin{equation}
    % \footnotesize
    \begin{aligned}
     |P^* - D^{*}_{\epsilon, N, \mathcal{E}_{train}}(\gamma)| \leqslant L\gamma +  \epsilon t (1 + \parallel \tau^*_{pert} \parallel_{1}) 
      + \mathcal{O}(\sqrt{\log(N)/N}),
    \end{aligned}
    \label{eq.optimazation_7}
\end{equation}
where L is the Lipschitz constant of $P^*(\gamma)$.
\end{theorem}
Assuming that both Assumption~\ref{assumption_1} and \ref{assumption_2} hold, the Problem~\ref{prob:problem1} is closely approximated by the duality gap in Eq.~(\ref{eq.optimazation_5}) when 1) the $\gamma$ is small, 2) $\mathcal{U}$ is a close approximation of $\mathcal{F}$, and 3) there are enough samples being observed.
We provide the proof of Theorem~\ref{theorem} in Appendix~\ref{proof}.
\section{Experiment Settings}
% In this study, we conduct extensive experiments to showcase the effectiveness and robustness of the proposed \sysname{} model.
% We compare the model's classification against popular competitive baselines on four real data scenarios.
% Additionally, we assess the universality of the model by employing different GNN backbones.
% we also conduct ablation studies on our model to demonstrate its effectiveness.

% \subsection{Experimental settings}
\textbf{Datasets.} In our experiments, we adopt four real network datasets, \textsc{Twitch-Explicit}~\cite{rozemberczki2021multi},  \textsc{Facebook-100}~\cite{traud2012social}, \textsc{WebKB}~\cite{pei2020geom} and \textsc{DBLP}~\cite{bojchevski2017deep}.
Each dataset contains a different number of graphs and more details are shown in Appendix~\ref{dataset}.

\textbf{Baselines.} We compare our model \sysname{} with standard empirical risk minimization (ERM), $3$ graph-specific methods (EERM~\cite{wu2022handling}, SRGNN~\cite{zhu2021shift}, and Mixup~\cite{zhang2017mixup}), and other $3$ OOD algorithms (IRM~\cite{arjovsky2019invariant}, DANN~\cite{ganin2016domain}, and VREx~\cite{krueger2021out}). More baselines details are presented in Appendix~\ref{baseline}.

\textbf{Settings for Domain Generalization.}
\label{setting}
% To assess the performance of \sysname{} on the testing graph with an unknown distribution, we train the model using multiple observed graphs and evaluate it on one or more unpredictable graphs. 
% In the same dataset, graphs share the same feature and label space, but they may have varying feature distributions and sizes.
% We utilize four real-world datasets: \textsc{Twitch-Explicit} (\textsc{Twitch}), \textsc{Facebook-100} (\textsc{Fb-100}), \textsc{WebKB}, and \textsc{DBLP}, all of which satisfy the aforementioned conditions.
We follow the principle of multiple graph training. 
In the \textsc{Twitch} dataset, we adopt a sequential approach where one out of the seven graphs is selected for testing, while the remaining six graphs are utilized for training and validation.
%In \textsc{Facebook-100}, following \cite{wu2022handling}, we use Penn, Brown, and Texas for testing, Cornell and Yale for validation, and use three different combinations from the remaining graphs for training, \textit{e.g.}, Hopkins + Caltech + Amherst, Bingham + Duke + Princeton and WashU + Brandeis+ Carnegie. 
In \textsc{Facebook-100}, following \cite{wu2022handling}, we designate Penn, Brown, and Texas for testing, employing three distinct combinations from the remaining graphs for training, Hopkins + Caltech + Amherst, Bingham + Duke + Princeton and WashU + Brandeis+ Carnegie. 
For \textsc{WebKB} and \textsc{DBLP}, we alternate the utilization of three graphs of the dataset as testing domains.
We utilize GCN as the backbone model and evaluate its performance using test accuracy and F1 score on four datasets.
In addition, the hyperparameter settings are provided in Appendix.~\ref{hyperpara}.

\section{Experimental Results}
\begin{figure*}
\centering
\begin{minipage}[b]{0.56\textwidth}
\centering
 \includegraphics[width=\textwidth]{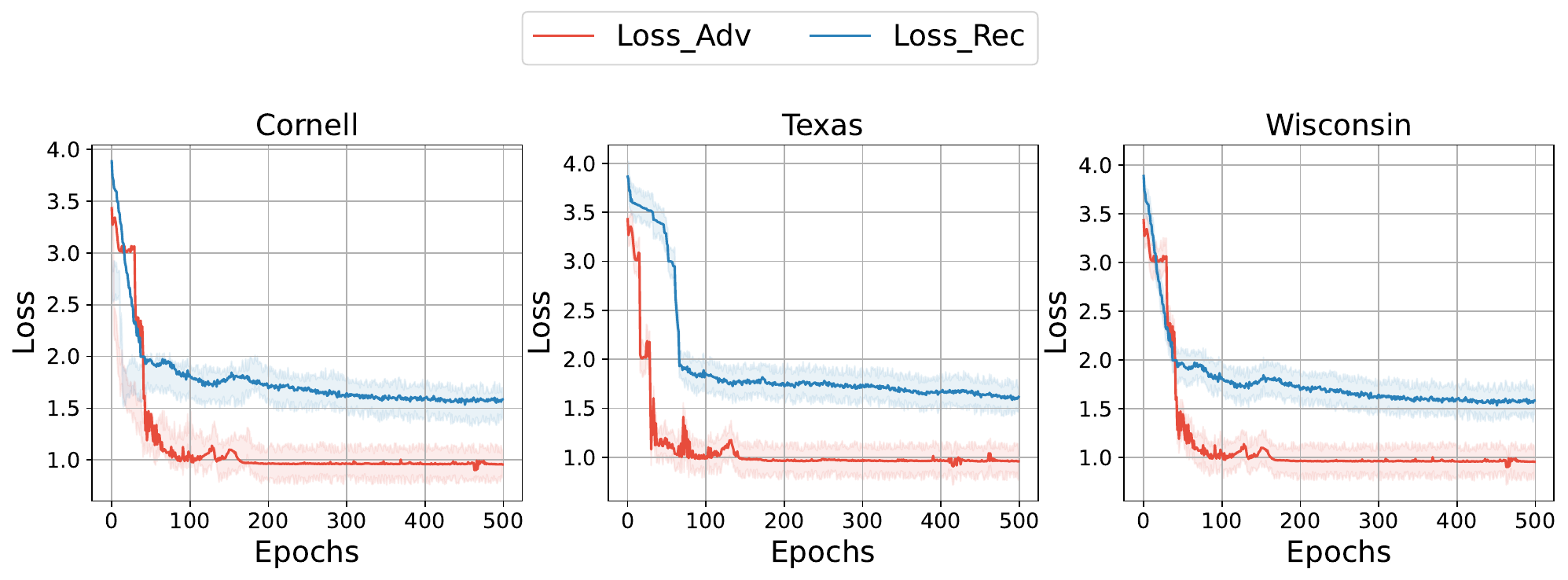}
 % \vspace{-3mm}
\caption{Loss curve of training discriminator where Loss\_Adv represents the adversarial loss and Loss\_Rec represents the reconstruction loss.}
\label{fig:loss}
\end{minipage}
\hfill
\begin{minipage}[b]{0.43\textwidth}
    \centering
    \includegraphics[width=\textwidth]{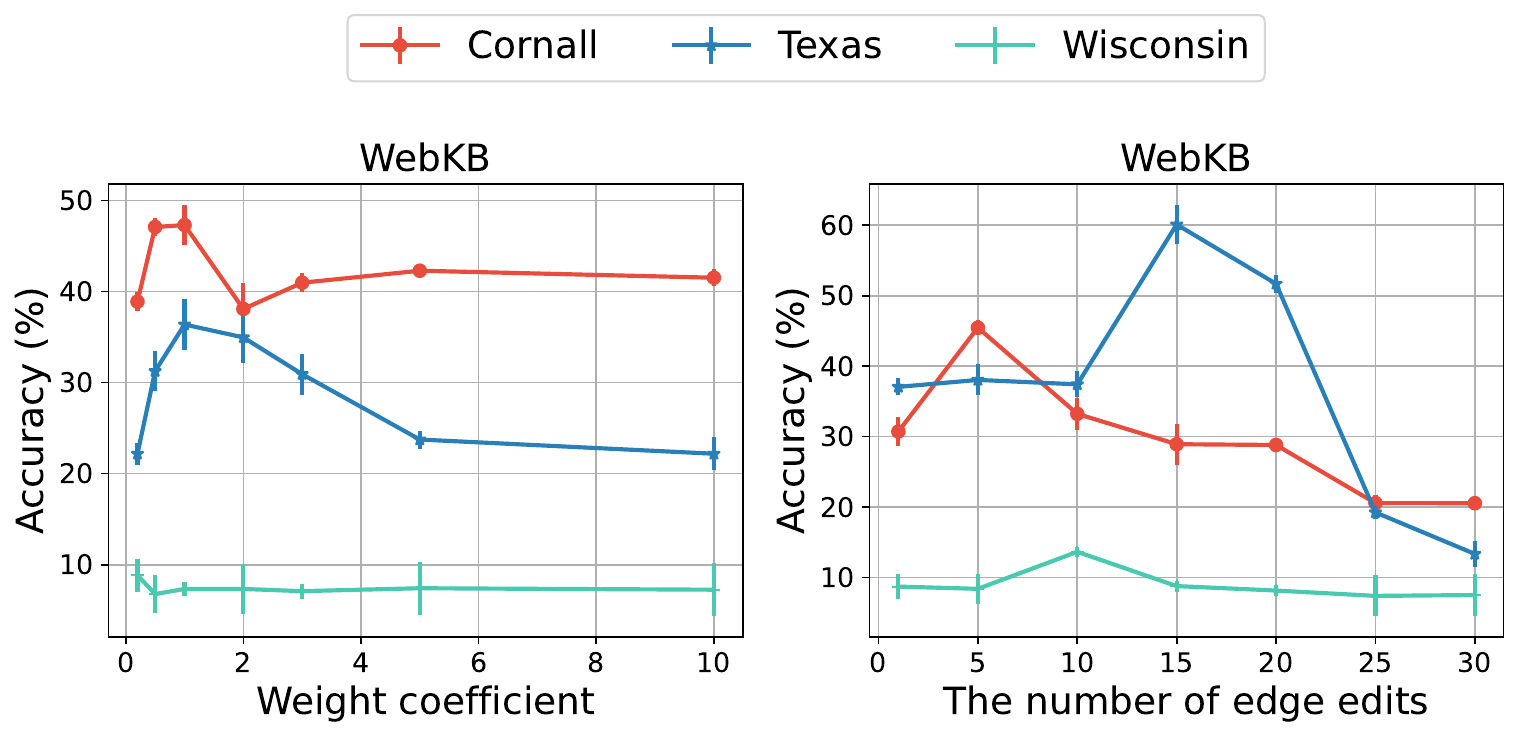}
    \vspace{-3mm}
    \caption{Sensitivity of \sysname{} w.r.t the weight coefficient in Eq.~(\ref{eq.invarant_3}) and the number of edits for each node by the edge.}
    \label{fig:test_a_s}
\end{minipage}
% \vspace{-3mm}
\end{figure*}

\textbf{Visualization and Analysis.} We conducted experiments to evaluate the performance of \sysname{} on node classification tasks.
% To present how effectively our model provably works on graph classification tasks. More visually and coherently, we visualize and analyze test results in detail.
%Because different results are presented on multiple graphs based on the same training datasets,  we finally compute the average value of different test data. 
The results are reported in Table~\ref{tab.acc_WebKB} - Table~\ref{tab.f1_Twitch}.
Due to space constraints, the results of accuracy on \textsc{Twitch} and the F1 score on \textsc{Twitch} and \textsc{FB-100} are shown in Appendix~\ref{experiment}.
% The table displays results for each graph used in testing, with the final dataset result in the last column, calculated as the average across all graphs. 
%Rows represent baseline method outcomes on different testing datasets. 
The training process is repeated $5$ times, and the resulting average is considered the final result.
As observed from the result, we have the following findings.

\begin{table}[t]
  \scriptsize
  \caption{Test accuracy (\%) comparison on \textsc{WebKB}.}
  % \vspace{-3mm}
  \label{tab.acc_WebKB}
  \centering
  \begin{tabular}{lccc|c}
    \toprule
    \textbf{Methods} & \textbf{Cornell}  & \textbf{Texas} & \textbf{Wisconsin}  & \textbf{Avg}\\
    \midrule
    EERM & $19.2 \pm 1.3$ & $31.1 \pm 2.1$ & $7.7 \pm 1.7$ & $19.4$ \\
    ERM & $12.8 \pm 2.0$ & \underline{$42.8 \pm 6.9$} & $13.5 \pm 1.2$ & $23.0$\\
    IRM & $15.3 \pm 0.7$ & $11.9 \pm 4.1$ & $13.4 \pm 0.9$ & $13.5$\\
    DANN & $15.8 \pm 2.2$ & $18.2 \pm 3.5$ & \underline{$14.1 \pm 1.1$} & $16.1$\\
    VREx & $15.3 \pm 1.0$ & $15.8 \pm 2.9$ & $14.0 \pm 1.7$ & $15.0$\\
    SRGNN& $15.8 \pm 0.6$ & $14.3 \pm 2.8$ & $8.7 \pm 2.5$ & $12.9$\\
    Mixup & $25.7 \pm 2.1$ & $20.6 \pm 1.0$ & $13.8 \pm 2.8$ & $20.0$ \\
    GroupDRO &  $15.8 \pm 1.3$ & $14.3 \pm 0.7$ & $13.0 \pm 1.5$ & $14.4$\\
    \midrule
    \sysname{}-C & $25.1 \pm 4.8$ & $30.6 \pm 5.4$ & $8.8 \pm 1.0$ & $21.5$ \\
    \sysname{}-A & \underline{$41.8 \pm 0.7$} & $32.8 \pm 0.4$ & $8.7 \pm 0.0$ & \underline{$27.8$} \\
    \sysname{} & $\mathbf{44.4 \pm 0.9}$ & $\mathbf{46.0 \pm 0.1}$ &  $\mathbf{14.1 \pm 1.6}$ & $\mathbf{34.8}$ \\
    \bottomrule
  \end{tabular}
% \vspace{-3mm}
\end{table}

\begin{table}[t]
\scriptsize
  \caption{Test accuracy (\%) comparison on \textsc{DBLP}.}
  % \vspace{-3mm}
  \label{tab.acc_DBLP}
  \centering
  \begin{tabular}{lccc|c}
    \toprule
    \textbf{Methods} & \textbf{D1}  & \textbf{D2} & \textbf{D3}  & \textbf{Avg}\\
    \midrule
    EERM & $43.3 \pm 3.7$ & \underline{$53.0 \pm 3.0$} & $49.6 \pm 3.1$ & $48.6 $ \\
    ERM & $38.3 \pm 2.4$ & $51.3 \pm 0.2$ & \underline{$50.0\pm 1.5$} & $46.5 $\\
    IRM & $35.6 \pm 0.7$ & $40.8 \pm 2.1$ & $43.4 \pm 0.2$ & $39.9 $\\
    DANN & $39.1 \pm 1.2$ & $42.8 \pm 1.4$ & $43.6 \pm 2.1$ & $41.8 $\\
    VREx& $43.5 \pm 0.4$ & $46.2 \pm 2.4$ & $45.9 \pm 0.2$ & $45.2$\\
    SRGNN& $44.7 \pm 0.8$ & $47.2 \pm 1.9$ & $47.0 \pm 2.6$ & $46.3 $\\
    Mixup & $45.9 \pm 1.7$ & $47.1 \pm 2.0$ & $48.3 \pm 0.9$ & $47.1 $ \\
    GroupDRO &  $46.2 \pm 1.5$ & $49.3 \pm 1.7$ & $48.4 \pm 0.3$ & $48.0$\\
    \midrule
    \sysname{}-C & \underline{$50.3 \pm 1.8$} & $50.9 \pm 0.2$ & $47.4 \pm 2.1$ & $49.5 $ \\
    \sysname{}-A & $50.2 \pm 0.3$ & $52.8 \pm 0.1$ & $49.5 \pm 1.0$ & \underline{$50.8$} \\
    \sysname{} & $\mathbf{51.9 \pm 1.0}$ & $\mathbf{53.2 \pm 2.6}$ & $\mathbf{50.1 \pm 1.1}$  & $\mathbf{51.8}$ \\
    \bottomrule
  \end{tabular}
% \vspace{-5mm}
\end{table}

\begin{table*}[t]
  \scriptsize
  \setlength\tabcolsep{3.5pt}
  \caption{Test accuracy (\%) on \textsc{Fb-100} where we compare different configurations of training graphs. }
  % \vspace{-3mm}
  \label{tab.acc_fb100}
  \centering
  \begin{tabular}{lcccl | cccc | ccc|c}
    \toprule
    ~ & \multicolumn{4}{c}{\textbf{Hopkins + Caltech + Amherst}} & \multicolumn{4}{|c|}{\textbf{Bingham + Duke + Princeton}} & \multicolumn{4}{c}{\textbf{WashU + Brandeis+ Carnegie}}\\
    \midrule
    \textbf{Methods}& \textbf{Penn}  & \textbf{Brown} & \textbf{Texas}  & \textbf{Avg} & \textbf{Penn}  & \textbf{Brown} & \textbf{Texas}  & \textbf{Avg} & \textbf{Penn}  & \textbf{Brown} & \textbf{Texas}  & \textbf{Avg}\\
    \midrule
    EERM &  \underline{$50.9 \pm 1.4 $}  & $55.4 \pm 4.1$ & $53.8\pm 4.6$ & $53.4$ & $50.7 \pm 2.2$ & $48.2 \pm 4.2$ & $50.6\pm 5.4$ & $49.8$ & \underline{$50.5 \pm 1.2$ }& $47.3 \pm 1.9$ & $51.2 \pm 4.1$ & $49.7$\\
    ERM & $50.1 \pm 1.5$ & $54.7 \pm 4.7$ & $55.1 \pm 2.4$ & $53.3$  & $47.3 \pm 0.5$ & $47.4 \pm 4.4$ & $50.5 \pm 4.6$ & $48.4$ & $49.8 \pm 0.6$ & $43.9 \pm 4.5$ & $47.6 \pm 4.1$ & $47.1$\\
    IRM & $45.3 \pm 0.2$ & $42.1\pm 1.2$ & $39.4 \pm 1.1$ & $42.3$ & $46.7\pm 0.1$ & $43.6\pm 2.2$ & $40.3 \pm 1.1$ & $43.5$  & $47.7 \pm 0.4$ & $44.8\pm 1.0$ & $42.3 \pm 2.8$ & $45.0$ \\
    DANN & $42.1 \pm 0.2$ & $43.4 \pm 2.1$ & $46.6 \pm 3.4$ & $44.1$ & $41.7 \pm 0.4$ & $42.8 \pm 1.1$ & $45.4 \pm 0.8$ & $43.3$ & $43.6 \pm 1.1$ & $44.8 \pm 2.2$ & $45.5\pm 1.3$ & $44.6$\\
    VREx& $41.5 \pm 2.4$ & $50.7 \pm 2.1$ & $47.9 \pm 2.1$ & $46.7$ & $42.0 \pm 0.7$ & $50.6 \pm 1.7$ & $46.9 \pm 1.9$ & $46.5$ & $45.6 \pm 1.2$ & $51.2 \pm 2.1$ & $45.9\pm 2.8$ & $47.6$ \\
    SRGNN & $40.3 \pm 1.7$ & $41.1 \pm 0.8$ & $46.4 \pm 1.0$ & $42.6$  & $39.9 \pm 0.9$ & $40.3 \pm 1.5$ & $44.9 \pm 2.1$ & $41.7 $ & $43.5 \pm 2.1$ & $42.3 \pm 0.9$ & $43.9 \pm 1.1$ & $43.3$\\
    Mixup & $50.6 \pm 2.1$ & $45.4 \pm 3.1$ & $48.9 \pm 3.1$ & $48.3$ & $50.5 \pm 2.0$ & $46.1\pm 0.8$ & $46.9 \pm 1.5$ & $47.9 $  & $50.0 \pm 1.9$ & $45.1\pm 1.6$ & $45.9\pm 2.3$ & $47.0$\\
   GroupDRO &  $48.2 \pm 2.4$ & $49.7 \pm 1.0$ & $50.0 \pm 1.3$ & $49.3 $ &  $47.1 \pm 1.3$ & $48.8 \pm 0.9$ & $49.9 \pm 1.6$ & $48.6 $ &  $46.8 \pm 0.1$ & $47.5 \pm 1.9$ & $48.2 \pm 2.1$ & $47.5$ \\
   \midrule
   \sysname{}-C & $49.9 \pm 0.6$ & $55.8 \pm 2.8$ & $55.2 \pm 3.2$ & $53.7 $ & $49.9 \pm 1.1$ & $54.4 \pm 2.2$ & $56.3 \pm 0.0$ & \underline{$53.6$} & $50.1 \pm 0.9$ & \underline{$56.8 \pm 2.2$} & \underline{$56.3 \pm 1.7$} & \underline{$54.5$} \\
   \sysname{}-A & $49.8 \pm 0.7$ & \underline{$56.0 \pm 2.1$} & \underline{$55.2 \pm 3.1$}  & \underline{$53.7$ } & \underline{$50.9 \pm 0.1$} & $\mathbf{56.9 \pm 0.0}$ &$\mathbf{56.4 \pm 0.0}$ &$\mathbf{54.7}$ & $50.0 \pm 0.7$ & $51.8 \pm 1.7$ &$51.5 \pm 2.5$ & $51.1$ \\
   \sysname{} & $\mathbf{50.9 \pm 0.8}$ & $\mathbf{58.0 \pm 0.3}$ & $\mathbf{56.3 \pm 0.0}$ & $\mathbf{55.1}$  & $\mathbf{50.8 \pm 0.0}$  & \underline{54.4 $\pm$ 3.6}& 
   \underline{53.9 $\pm$ 3.2} & 53.1& $\mathbf{50.5 \pm 0.7}$ & $\mathbf{56.9 \pm 2.3}$ & $\mathbf{56.3 \pm 0.0}$& $\mathbf{54.6}$\\
    \bottomrule
  \end{tabular}
% \vspace{-5mm}
\end{table*}

Firstly, \sysname{} outperforms current state-of-the-art methods. EERM ranks second only to our model for multi-graph generalization in Table~\ref{tab.acc_DBLP}, Table~\ref{tab.acc_fb100}, and Table~\ref{tab.acc_Twitch}.
In contrast, EERM does not achieve the second-best performance in Table~\ref{tab.acc_WebKB}. 
This is because \textsc{WebKB} is very sparse, and the EERM generates domains by randomly adding and subtracting edges. This introduces significant changes to the graph, thereby increasing the difficulty of invariant learning.
We generate diverse domains through attribute matrix transition and topology augment, aiming to maximize graph variations while preserving semantic information.
The augmentation applied in both attribute and topology structures in \sysname{} is deemed more realistic.

Secondly, the compared approaches show varied performances amidst distinct distribution shifts.
Traditional invariant learning algorithms such as IRM and VREx exhibit lower performance compared to graph data augmentation methods. 
This discrepancy arises from the strong dependence of invariant learning on the environment. 
The former predominantly relies on real multi-graph inputs to extract invariant information, whereas the latter can construct environments with maximum variability to enhance model generalization.
Compared with the general OOD method, the graph-based OOD method has better performance, which shows that graph-based OOD is more effective for classifying tasks on graphs.

\begin{figure}
  \centering
  \includegraphics[width=0.5\linewidth]{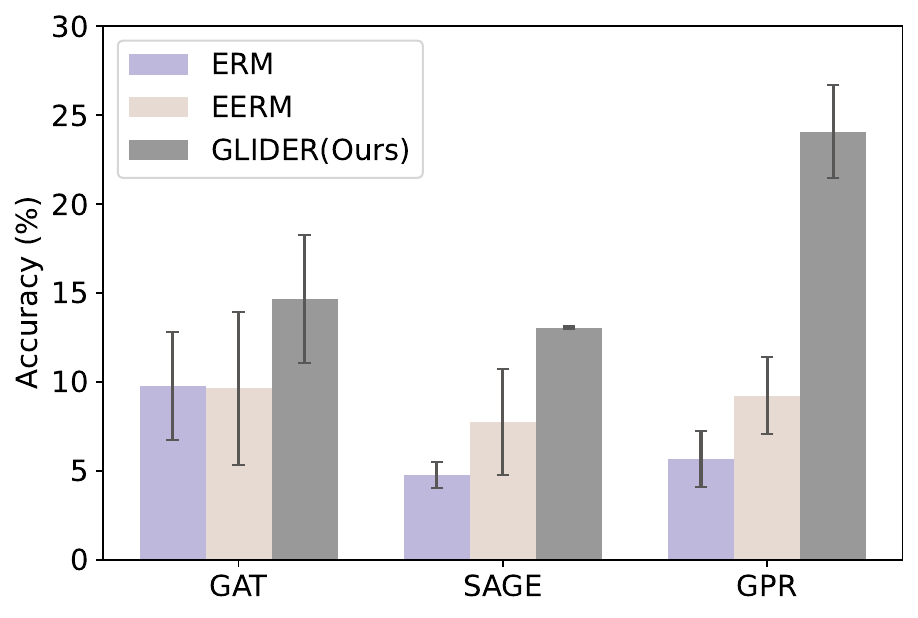}
  \vspace{-3mm}
  \caption{The accuracy of \sysname{} (Ours) and another two baseline methods using different GNN backbones on \textsc{WebKB}.}
  \label{fig:backbone}
  % \vspace{-5mm}
\end{figure}

% \begin{figure}[htbp]
% 	\centering
% 	\begin{subfigure}{0.32\linewidth}
% 		\centering
% 		\includegraphics[width=\linewidth]{figures/loss_0.pdf}
% 		\caption{Training: Texas + Wisconsin}
% 		\label{loss_0}
% 	\end{subfigure}
% 	\centering
% 	\begin{subfigure}{0.32\linewidth}
% 		\centering
% 		\includegraphics[width=\linewidth]{figures/loss_1.pdf}
% 		\caption{Training: Cornell + Texa}
% 		\label{loss_1}
% 	\end{subfigure}
% 	\centering
% 	\begin{subfigure}{0.32\linewidth}
% 		\centering
% 		\includegraphics[width=\linewidth]{figures/loss_2.pdf}
% 		\caption{Training: Cornell + Wisconsin}
% 		\label{loss_2}
% 	\end{subfigure}
% 	\caption{ The change of losses as epochs increase on \textsc{WebKB}. Loss\_Adv represents the adversarial loss and Loss\_Rec represents the reconstruction loss.}
% 	\label{Fig.loss}
%  %\vspace{-5mm}
% \end{figure}

In addition, we evaluate the impact of various GNN backbones on the framework.
We separately utilize GAT, SAGE, and GPR as the GNN models on \textsc{WebKB}.
%The layer numbers are set as 2 for all GNN models, the hidden size is set as $32$ and the head number is set as 2 for GAT.
The accuracy and F1 score results above demonstrate that EERM and ERM outperform other baseline methods.
Therefore, we solely compare the performance of EERM and ERM with \sysname{} using different backbones on \textsc{WebKB} and report the results in Fig.~\ref{fig:backbone}.
We observed that our model significantly outperforms EERM and ERM, with a maximum improvement of $13.0\%$ in accuracy.
This phenomenon demonstrates the universality and robustness of our model.

We also illustrate how the loss of the first module, attribute matrix transformation, changes as the epoch increases in Fig.~\ref{fig:loss}.
The loss of the first stage contains the adversarial loss and the reconstruction loss, both of which converge to stop the training process.
In conclusion, \sysname{} maintains the superiority of training graphs having distribution shifts, thus affirming the effectiveness of our model in addressing node-level OOD generation.
% The above results demonstrate its power on various real-world datasets for handling distribution shifts.

\textbf{Ablation Study.} To investigate the contribution of each module in \sysname{}, we individually remove two key
modules, namely attribute matrix transformation and topology augmentation.
% we conduct two ablation studies to assess the impact of the attribute matrix transformation and topology augmentation modules.
% In this study, we continue to use test accuracy and F1 score as evaluation metrics, and we employ GCN as the GNN model to conduct experiments on four real-world datasets.
% \textbf{Attribute matrix transformation.} 
% The aim of attribute matrix transformation is to generate variable node attribute distribution.
% \textbf{Removing attribute matrix transformation.}
(1) We remain the semantic feature reconstruction $\mathcal{L}_{rec}^{c}$ and remove the other part, which we define as \sysname{}-C.
% \sysname{}-C represents weight coefficient
% $\lambda$, $\lambda_{x}$ and $\lambda_{s}$ are evenly set as 0 in \eqref{eq.totalLoss}.
%\textbf{Topology argumentation.} 
% Topology augmentation is dedicated to generating multiple subgraphs with diverse topologies.
% \textbf{Removing topology augmentation.}
(2) We generate multiple graphs with attribute distribution shifts while keeping the topology structures identical. 
We refer to this variant as \sysname{}-A.

The results are shown in Table~\ref{tab.acc_WebKB} - Table~\ref{tab.f1_Twitch}. 
Due to limited space, some results are available in Appendix~\ref{experiment}.
We can see that: (1) the performance of \sysname{}-C is slightly inferior to \sysname{} on all datasets. 
Compared with EERM which does not perform the transformation on the attribute distribution, the performance of \sysname{}-C is still elevated.
This phenomenon shows that the reconstruction of domain-specific variation factor also plays a prominent role, and (2) Compared with EERM, which does not generate multiple environments using topology structure augmentation, \sysname{}-A exhibits superior performance.
This indicates that minimizing the prediction variance of multiple input graphs with the largest possible difference in topology is effective for addressing the OOD problem.
% In summary, the approach that integrates topology and attribute matrix transition to generate multiple graphs and then learns the invariant representation by minimizing the empirical risk from multiple virtual environments performs better for addressing the OOD problem.

% \begin{figure}[htbp]
% 	\centering
% 	\begin{subfigure}{0.4\linewidth}
% 		\centering
% 		\includegraphics[width=\linewidth]{figures/a_test.pdf}
% 		\caption{$\alpha$ denotes the weight coefficient of mean loss in \eqref{eq.invarant_3}}
% 		\label{fig:a_test}
% 	\end{subfigure}
% 	\centering
% 	\begin{subfigure}{0.4\linewidth}
% 		\centering
% 		\includegraphics[width=\linewidth]{figures/s_test.pdf}
% 		\caption{$s$ denotes the number of edge edits for each node.}
% 		\label{fig:s_test}
% 	\end{subfigure}
% 	\caption{ ACC for hyper parameter $\alpha$ and $s$ on \textsc{WebKB}.}
% 	\label{Fig.para}
%  \vspace{-2mm}
% \end{figure}

\textbf{Sensitivity Analysis.} We evaluated the sensitivity of our model regarding the $\alpha$ and $s$ parameters on \textsc{WebKB} and show the result in Fig.~\ref{fig:test_a_s}. 
$\alpha$ is the weight for combination, while $s$ is the number of editing per node by edge.

%We set $\alpha$ in $[0.2, 0.5, 1, 2, 3, 5, 10]$.
% For Wisconsin, the accuracy value does not change significantly as $\alpha$ increases.
% For both Cornell and Texas, the optimal value is achieved when $\alpha=1$, indicating that the optimal accuracy can be obtained on these two datasets while maintaining the same variance and mean of risks from various generated graph weights.
Intuitively, we observe that as the value of $\alpha$ changes, the accuracy on all three graphs tends to stabilize when $\alpha$ is greater than $4$.
Therefore, \sysname{} is robust to parameters $\alpha$. 
The variation of accuracy values in the three graphs tends to stabilize with the increase of $s$, and we determine the value of $s$ that yields the highest accuracy as the optimal model parameter.
Specifically, the optimal $s$ values differ for the three graphs. 
% When $s=5$, Cornell achieves the best performance, when $s=10$, the optimal value is obtained for Wisconsin, and when $s=15$, the best performance is achieved in Texas.
The parameter $s$ exhibits varying degrees of influence on different graphs, likely due to the discrepancy in network density among the graphs.

\section{Conclusion}
In this paper, we propose an effective framework \sysname{} for node-level OOD generation on graphs.
% The node-level OOD generation on graphs aims to enhance the model’s generalization ability for node label prediction across domains.
% However, capturing invariant information from different domains to achieve stable prediction across domains can be challenging due to the distribution shifts of nodes in graphs occurring on both the attribute and topology.
%In terms of this issue, we diversify the variation by modeling the potential seen or unseen variations of attribute distribution and topology of nodes and then minimize the semantic gap for each class among these variations.
We diversify the variation by modeling the potential seen or unseen variations of attribute distribution and topology of nodes and then minimize the semantic gap for each class among these variations.
First, \sysname{} generates multiple domains by attribute matrix transition and topology augmentation, and the distributions of node attributes and topology vary across domains.
%Because it is impractical to consider each node in graph-structured data independently, we introduce the ego-graph for each node to capture the surrounding influences on the centered node. 
Then, we train a domain-invariant classifier by minimizing the empirical risk from multiple domains to learn the invariant information of nodes. 
Finally, we construct abundant experiments to evaluate the effectiveness of our framework in the node classification task, and the experimental results show that our method is superior to some baseline methods.

\section*{Acknowledgments}
% This was was supported in part by......

%Bibliography
\bibliographystyle{unsrt}  
\bibliography{references}

\appendix
\clearpage
% \onecolumn
\section{Appendix}
% \subsection{Experimental Code}
% Code is available for review purposes at: \\ \href{https://anonymous.4open.science/r/Node-IDGG-3DF1}
% {https://anonymous.4open.science/r/Node-IDGG-3DF1}

\subsection{Notations}
\label{notaions}
% \vspace{-6mm}
In this paper, vectors are denoted by lowercase boldface letters, \textit{e.g.,} $\mathbf{x}$. 
Scalars are denoted by lowercase italic letters, \textit{e.g.,} $\alpha$ and $\lambda$. 
Matrices are denoted by capital italic letters, \textit{e.g.,} $A$. 
Sets are denoted by uppercase calligraphic letters, \textit{e.g.,} $\mathcal{E}$.
%All the random variables are denoted as bold uppercase letters, \textit{e.g.,} $\mathbf{G}$.
All notations used in this paper are listed in Table~\ref{tab.notations}.

% \vspace{3mm}
\begin{table}[htbp]
\caption{Important notations and their description.}
% \vspace{-3mm}
\begin{center}
\begin{tabular}{c|p{6.3cm}}
\toprule
\textbf{Notation} & \textbf{Description}\\
\midrule
$G$ & An input graph \\
$A$ & The adjacent matrix of a graph $G$\\
$X$ & The attribute feature matrix of a graph $G$\\
$G^e$ & A graph sampled from domain $e$ \\
$A^e$ & The adjacent matrix of a graph $G^e$\\
$X^e$ & The attribute feature matrix of a graph $G^e$\\
$\mathcal{V}$ & The node set of a graph $G$ \\
$|\mathcal{V}|$ & The node number of a graph $G$ \\
$\mathbf{y}$ & The label of a graph $G$ \\
$\mathcal{G}$ & An input graph space \\
% $\mathcal{Y}$ & An label space \\
$\mathcal{E}$ & The set of domain $e$ \\
$G_v^e$ & The ego-graph of node $v$ in $G^e$ \\
$y_v^e$ & The label of node $v$ in $G^e$ \\
$A_v^e$ & The adjacent matrix of $G_v^e$ \\
$X_v^e$ & The attribute feature matrix of $G_v^e$ \\
$f_c$ & A downstream classifier \\
%$\ell$ &  A loss function \\
$\mathbf{c}$ & A latent semantic factor \\
$\mathbf{r}^e$ & A latent variation factor from domain $e$ \\
$\mathbf{\hat{r}}_a$ & The variation factor of attributes draw from a prior distribution   \\
$\mathbf{r}_a^e$ & The variation factor on node attributes from domain $e$ \\
$G'$ & A graph where the attribute distribution of each node is shifted from the input graph $G$ \\
$G''$ & A graph where the attribute distribution and topology of each node are both shifted from the input graph \\
$A'$ & A newly generated adjacency matrix\\
$X'$  & A generated attribute matrix with distribution shifts \\
$X_c$ & The latent semantic factor on the attribute of a graph \\ 
$R_a$ & The latent variation factor on the attribute of a graph \\ 

$E^c$ & A semantic encoder \\
$E^r$ & A variation encoder \\
$D$ & A decoder \\
$\Psi$ &  A discriminator\\
%$h^*$ & An encoder \\
$l_g$, $l_f$ & The learning rates\\
$\lambda_x, \lambda_c, \lambda_x, \alpha $ & Weight coefficient for different loss \\
$T$ & The number of iterations in topology augmentation \\
$s$ & The number of edge editing actions for each node \\
\bottomrule
\end{tabular}
\label{tab.notations}
\end{center}
\end{table}

\vspace{3mm}

% \subsection{Algorithms}
% \label{algorithm}

% \begin{algorithm}[t]
%     \caption{\textbf{The optimize process of graph generators}} \label{gw_algorithm}
%     \begin{algorithmic}[1]
%     \STATE \textbf{Input:} generated  $G'=(A, X')$ and label $\mathbf{y'}$, parameters $\boldsymbol{\theta}$ of GNN, 
%     learning rates $l_g$, $l_f$.
%     \STATE \textbf{Initialize} parameters of graph editers $\{w_k\}_{k=1}^K$
%     %\STATE \textbf{Repeat}: 
%     \STATE \textbf{For} $t=1$, $\cdots$, $T$ \textbf{do}
%     \STATE \qquad Generated graphs $G^k = (A_k, X')$ from graph \\ \ \  \  \ \ \ \ \ generator $g_{w_k}, k=1, \cdots, K$
%     %\STATE \qquad \qquad Compute loss $\mathcal{L}(w) = \mathbb{V}(l_k)$
%     \STATE  \qquad Compute loss $\mathcal{L}(w) = \mathbb{V}(l_k)$ in \eqref{eq.gwk}
%     \STATE  \qquad Update $w_k \leftarrow w_k + l_g\bigtriangledown_{w_k}\log p_{w_k}(A_k) \mathcal{L}(w)$
%     \STATE \textbf{End for}
%     %\STATE \textbf{Until} convergence
      
%     \end{algorithmic}
% \end{algorithm}

% \clearpage
\section{Datasets And Experimental Details}
\subsection{Dataset}
\label{dataset}
The statistical characteristics of these networks are shown in Table~\ref{tab.data}. 
\vspace{3mm}

\begin{table}[htbp]
 \footnotesize
  \setlength\tabcolsep{2pt}
  \caption{Key Characteristics of Datasets }
  \label{tab.data}
  \centering
  \begin{tabular}{lccccc}
    \toprule
    \textbf{Dataset} &  \textbf{\# Nodes} & \textbf{\# Edges} & \textbf{\# Class} & \textbf{\#Feat} & \textbf{\# Domain}\\
    \midrule
    \textsc{Twitch-Explicit} &  20945 & 153,138 & 2 & 3170 & 7\\
    \textsc{Facebook-100} & 131,924 & 1,590,655  & 2  & 12412  & 12\\
    \textsc{WebKB} & 617 & 1138 & 5 &  1703 &3 \\
    \textsc{DBLP} &  17,716 & 105,734 & 4 & 1639 & 3\\
    \bottomrule
  \end{tabular}
\end{table}

\begin{itemize}[leftmargin=*]
    \item \textsc{Twitch-Explicit}~\cite{rozemberczki2021multi}. It is a gamer network that includes seven networks: DE, ENGB, ES, FR, PTBR, RU, and TW.
    Each network represents a particular game region. 
    The aforementioned networks have comparable sizes but vary in terms of densities and maximum node degrees. 
    %We take one of the seven for testing in turn and the rest for training and validation.
    
    \item \textsc{Facebook-100}~\cite{traud2012social}. This dataset comprises 100 snapshots of the \textit{Facebook} friendship network, dating back to 2005.
    Each node represents a user from a particular American university, and the edges indicate the friendships between these users.
    We use twelve networks in our experiments: John Hopkins, Caltech, Amherst, Bingham, Duke, Princeton, WashU, Brandeis, Carnegie, Penn, Brown, and Texas.
    %And select some of them for training.
    
    \item \textsc{WebKB}~\cite{pei2020geom}. It is a web page network dataset. 
    The nodes in the network represent web pages, and the edges symbolize hyperlinks connecting these web pages. Additionally, the node features are represented using the bag-of-words representation of the web pages.
    The task is to classify the nodes into one of five categories: student, project, course, staff, and faculty. We split \textsc{\textsc{WebKB}} into three networks (Cornell, Texas, and Wisconsin) according to the domain university. %Analogously, we take one of the three for testing respectively and the rest for training and validation.
    \item \textsc{DBLP}~\cite{bojchevski2017deep}. It is a citation network. 
    Nodes represent publications and edges represent citation links between these publications.
    The task involves classifying publication types into $4$ distinct classes.
    We generate three splits based on a domain selection (D1, D2, D3).
    The Domain is defined by the selected-word-count of a publication.
    %Correspondingly, one of the three splits is used for testing, while the remaining splits are utilized for training and validation purposes.  
\end{itemize}

\subsection{Compared Methods}
\label{baseline}
\begin{itemize}[leftmargin=*]
\item \textbf{EERM}~\cite{wu2022handling} generates environments with diverse topologies and then minimizes the variances and mean values of predicted loss across different environments.

\item \textbf{IRM}~\cite{arjovsky2019invariant} aims to discover a latent invariant representation that remains consistent across all environments. This is achieved by penalizing feature distributions that exhibit varying optimal linear classifiers for each environment.

\item \textbf{DANN}~\cite{ganin2016domain} learns to identify features that are indistinguishable between the training (source) and test (target) domains through adversarial training. 
%This is achieved by training both a regular classifier and a domain classifier.

\item \textbf{VREx}~\cite{krueger2021out} solves the OOD problem by minimizing the variance of training risks based on the Risk Extrapolation (REx) model.

\item \textbf{SRGNN}~\cite{zhu2021shift} is devoted to solving the distributional shift problem by converting biased data sets to unbiased data distribution.
This is achieved using a central moment discrepancy regularizer and the kernel mean matching technique.

\item \textbf{Mixup}~\cite{zhang2017mixup} improves model generation capacity by constructing novel training examples drawn from raw data, thereby expanding the training distribution.

\item \textbf{GroupDRO}~\cite{sagawa2019distributionally} minimizes the worst-case training loss in all environments by increasing regularization on DRO groups.

\end{itemize}

\subsection{Hyperparameter Tuning} 
\label{hyperpara}
The first module, attribute matrix transformation, include four hyperparameters:$\lambda_x$, $\lambda_c$, and $\lambda_s$.
They are all weight coefficients of different parts in Eq.~\eqref{eq.totalLoss}.

The parameters of the second module, topology augmentation, include four hyperparameters: $\alpha$ which denotes the weight coefficient of different parts in Eq.~\eqref{eq.invarant_3}, $K$ which denotes the number of generated adjacency matrices, $T$ which denotes the number of iterations, and $s$ which denotes the number of edge editing for each node.

We set $\alpha \in \{ 0.2, 0.5, 1.0, 2.0, 3.0, 5.0, 10.0\}$, $K \in \{1, 2, 3 \}$,
the learning rate $l_g, l_f \in \{ 0.0001, 0.0002, 0.001, 0.005, 0.01 \}$, $T \in \{1,5\}$, $s \in \{1, 5, 10, 15, 20, 25, 30\}$, weight coefficient $\lambda$, $\lambda_c$, $\lambda_s$ and $\lambda_x$ are all in the range $[0,1]$.

\section{Proofs}
\subsection{Sketch Proof of Theorem~\ref{theorem}}
\label{proof}
The Theorem~\ref{theorem}. Given $\epsilon > 0$ and let $\varphi$ be an $\epsilon$-parameterization of $\mathcal{F}$. Let $l$ and $d$ are under $[0,H]$ and both non-negative, convex.
while assume that $d(\mathbb{P},\mathbb{T})=0$ if and only if $\mathbb{P}=\mathbb{T}$.
Then assuming that $\mathcal{U}$ has finite VC-dimension with probability $1-\rho$ over the $N$ samples from $\mathbb{P}$ that
\begin{equation}
    % \footnotesize
    \begin{aligned}
     |P^* - D^{*}_{\epsilon, N, \mathcal{E}_{train}}(\gamma)| \leqslant &L\gamma +  \epsilon t (1 + \parallel \tau^*_{pert} \parallel_{1}) 
      + \mathcal{O}(\sqrt{\log(N)/N}),
    \end{aligned}
    % \label{eq.optimazation_7}
\end{equation}
% \textit{Proof.} 
% To prove Theorem~\ref{theorem} we first make the following assumption:
% \begin{assumption}
% \label{assumption_appendix}
% \end{assumption}
\begin{proof}
First, let us assume ${p}^*(\gamma)$ is L-Lipschitz continuous in $\gamma$, when $d(\mathbb{P},\mathbb{T})=0$ if and only if $\mathbb{P}=\mathbb{T}$ and ${P}^* = {P}^*(0)$, it follows that 
\begin{equation}
    % \footnotesize
    \begin{aligned}
     |P^* - P^*(\gamma)| & = |P^*(0) - P^*(\gamma)|\\
     & \leq L|0-\gamma| \\
     & = L\gamma
    \end{aligned}
    \label{eq.appendex_1}
\end{equation}
Second, given $\gamma >0$ and $l$ and $d$ are under $[0,H]$ and both non-negative, convex, we obtain that 
\begin{equation}
    % \footnotesize
    \begin{aligned}
     P^*(\gamma) \leqslant D^*_{\epsilon}(\gamma) \leqslant P^*(\gamma)+\epsilon t (1 + \parallel \tau^*_{pert} \parallel_{1}) \cdot \max\{L_l,L_d\}
    \end{aligned}
    \label{eq.appendix_2}
\end{equation}
where $\tau^*_{pert}$ is the optimal dual variable for a perturbed version of Eq.~\eqref{eq.optimazation_2} in which the constraints are tightened
to hold with margin $\gamma-\epsilon \cdot  \max\{L_l,L_d\}$. Then we get that
\begin{equation}
    % \footnotesize
    \begin{aligned}
     P^*(\gamma) - D^*_{\epsilon}(\gamma) \leqslant \epsilon t (1 + \parallel \tau^*_{pert} \parallel_{1}) \cdot \max\{L_l,L_d\}
    \end{aligned}
    \label{eq.appendix_3}
\end{equation}
Next, assuming that $l$ and $d$ are non-negative and bounded in $[-H,H]$ and $\mathcal{U}$ has finite VC-dimension with probability $1-\rho$ over the $N$ samples from $\mathbb{P}$ that
%let Assumption~\ref{assumption_appendix} hold and $
\begin{equation}
    % \footnotesize
    \begin{aligned}
     |D^*_{\epsilon} - D^{*}_{\epsilon, N, \mathcal{E}_{train}}(\gamma)| \leqslant  2H\sqrt{\frac{1}{N}\big[1+\log(\frac{4(2N)^{d_{VC}}}{\rho})\big]},
    \end{aligned}
    \label{eq.appendix_4}
\end{equation}
By combining Eq.~{\eqref{eq.appendex_1}}, Eq.~\eqref{eq.appendix_3} and Eq.~\eqref{eq.appendix_4}, we obtain that
\begin{equation*}
    % \footnotesize
    \begin{aligned}
     |P^* - D^{*}_{\epsilon, N, \mathcal{E}_{train}}(\gamma)| 
     &= |P^*+P^*(\gamma) - P^*(\gamma) + D^*_{\epsilon}(\gamma) - D^*_{\epsilon}(\gamma) -D^{*}_{\epsilon, N, \mathcal{E}_{train}}(\gamma)| \\
     &\leqslant |P^*-P^*(\gamma)| + |P^*(\gamma)-D^*_{\epsilon}(\gamma)| +|D^*(\gamma) -D^{*}_{\epsilon, N, \mathcal{E}_{train}}(\gamma)| \\
     &\leqslant L\gamma +\epsilon t (1 + \parallel \tau^*_{pert} \parallel_{1}) + 2H\sqrt{\frac{1}{N}\big[1+\log(\frac{4(2N)^{d_{VC}}}{\rho})\big]}
    \end{aligned}
    \label{eq.appendix_5}
\end{equation*}
\end{proof}

% \begin{theorem}[Data-dependent duality gap]
% \label{theorem}
% Given $\epsilon > 0$ and let $\varphi$ be an $\epsilon$-parameterization of $\mathcal{F}$.
% Under mild regularity assumptions on $\ell$ and $h$ and assuming that $\mathcal{Z}$ has finite VC-dimension, with probability $1-\rho$ over the $N$ samples from $\mathbb{P}(\bar{{X}}, {Y})$, we have that
% \begin{equation}
%     \footnotesize
%     \begin{aligned}
%      |P^* - D^{*}_{\epsilon, N, \mathcal{E}_{train}}(\gamma)| \leqslant &L\gamma +  \epsilon t (1 + \parallel \tau^*_{pert} \parallel_{1}) \\
%      & + \mathcal{O}(\sqrt{\log(N)/N}),
%     \end{aligned}
%     \label{eq.optimazation_7}
% \end{equation}
% where L is the Lipschitz constant of $P^*(\gamma)$.
% \end{theorem}

\subsection{More Experimental Results}
\label{experiment}
To further assess the classification performance of our model, we also evaluate the F1 score on \textsc{Twitch} and \textsc{Fb-100}, both of which contain two types of labels.
In Table~\ref{tab.f1_fb100} and Table~\ref{tab.f1_Twitch}, we show the results for the F1 score on \textsc{Twitch} and \textsc{Fb-100}.
In comparison to all baselines, our approach exhibits a significant improvement in average F1 scores.
From a comprehensive perspective, the EERM and the GroupDRO both rank second only to ours in terms of classification performance. 
Moreover, they exhibit favorable performance compared to other methods across multiple datasets.
\begin{table*}[t!]
  \footnotesize
  % \setlength\tabcolsep{2pt}
 %\fontsize{12pt}{18pt}\selectfont
  \caption{Test accuracy (\%) on \textsc{Twitch} where we compare different configurations of training graphs. }
  \label{tab.acc_Twitch}
  \centering
  \begin{tabular}{lccccccc|c}
    \toprule
   \textbf{Methods} & \textbf{ENGB}  & \textbf{ES} & \textbf{DE} & \textbf{FR} & \textbf{PTBR} & \textbf{RU} & \textbf{TW} & \textbf{Avg}\\
    \midrule
    EERM & $45.7 \pm 0.2$ & \underline{70.7 $\pm$ 0.2} & $50.0 \pm 1.8$ & $58.2 \pm 4.4$ & $65.3 \pm 0.0$ & $72.7 \pm 3.8$ & \underline{$60.7 \pm 0.0$} & $60.5$ \\
    ERM & $45.6 \pm 0.0$ & $70.6 \pm 0.1$ & $40.2 \pm1.3$ & \underline{$63.0 \pm 0.1$} & $65.0 \pm 0.1$ & $66.3 \pm 8.5$ & $58.9 \pm 2.9$ & $58.5$ \\
    IRM & $42.4 \pm 0.4$ & $63.1 \pm 2.0$ & $39.0 \pm 0.9$ & $49.9 \pm 0.1$ & $51.6 \pm 0.9$ & $55.3 \pm 1.5$ & $46.0 \pm 0.7$ & $49.6$ \\
    DANN & \underline{$46.5 \pm 1$} & $49.5 \pm 1.1$ & $36.1 \pm 2.1$ & $42.2 \pm 1.1$ & $51.8 \pm 3.0$ & $48.7\pm 1.6$ & $49.7 \pm 1.5$ & $46.6 $\\
    VREx & $39.7 \pm 1.0$ & $33.3 \pm 4.0$ & $49.2 \pm 1.1$ & $47.5 \pm 0.8$ & $49.8 \pm 3.1$ & $58.7 \pm 0.1$ & $49.5 \pm 1.4$ & $46.8 $\\
    SRGNN& $40.3 \pm 2.0$ & $53.8 \pm 1.5$ & $44.9 \pm 0.1$ & $48.4 \pm 0.4$ & $38.8\pm 2.6$ & $58.8 \pm 1.1$ & $44.3 \pm 3.7$ & $47.0 $\\
    Mixup & $37.2 \pm 1.0$ & $48.7 \pm 2.4$ & $48.4 \pm 1.5$ & $39.9 \pm 2.9$ & $46.9\pm 1.8$ & $59.1 \pm 0.8$ & $56.6 \pm 0.9$ & $48.1$\\
   GroupDRO & $45.1 \pm 0.9$ & $55.6 \pm 0.4$ & \underline{$51.0 \pm 1.1$ } & $53.1 \pm 1.2$ & $51.1\pm 2.0$ & $67.1 \pm 2.1$ & $47.3 \pm 0.4$ & $53.1$\\
   \midrule
   \sysname{} -C & $45.9 \pm 0.4$ & $69.9 \pm 0.4$ & $50.2 \pm 3.1$ &$59.9 \pm 3.8$ & $65.0 \pm 3.7$ & \underline{$74.2 \pm 2.1$} & $59.6 \pm 2.0$ & \underline{$60.7$} \\
   \sysname{} -A & $45.4 \pm 0.0$ &$70.7 \pm 0.0$ & $47.0 \pm 2.4$ & $61.9 \pm 2.2$ & \underline{$65.4 \pm 0.0$} & $72.0 \pm 2.2$ & $58.4 \pm 4.9$ & $60.1$ \\
    \sysname{} & \textbf{46.7}$\pm$ \textbf{0.6} & \textbf{70.7} $\pm$ \textbf {0.7} & \textbf{51.1}$\pm$ \textbf{2.8} & \textbf{63.1} $\pm$ \textbf{1.2} & \textbf{65.4} $\pm$ \textbf{1.2} & \textbf{75.5} $\pm$ \textbf{1.2}  & \textbf{60.8} $\pm$ \textbf{2.8}& $ \textbf{61.9} $ \\
    \bottomrule
  \end{tabular}
\end{table*}

\begin{table*}[t]
 \footnotesize
  \caption{F1 score (\%) on \textsc{Twitch} where we compare different configurations of training graphs. }
  \label{tab.f1_Twitch}
  \centering
  \begin{tabular}{lccccccc|c}
    \toprule
    \textbf{Methods} & \textbf{ENGB}  & \textbf{ES} & \textbf{DE} & \textbf{FR} & \textbf{PTBR} & \textbf{RU} & \textbf{TW} & \textbf{Avg}\\
    \midrule
    EERM & $34.6 \pm 2.0$ & $60.4 \pm 2.2$ & \textbf{53.2} $\pm$ \textbf{1.9} & $51.0 \pm 2.0$ & $52.0 \pm 0.4$ & \textbf{65.9} $\pm $ \textbf{0.7} & $49.2\pm 2.1$ & $52.3 $ \\
    ERM & $40.8 \pm 0.1$ & $59.9 \pm 0.0$ & $40.1 \pm 0.9$ & \textbf{57.2} $\pm$ \textbf{0.0} & $53.8 \pm 2.1$ & $54.1 \pm 3.1$ & $45.2 \pm 1.6$ & $50.1 $ \\
    IRM & $39.9 \pm 2.2$ &  \underline{$61.0 \pm 0.5$} & $38.9 \pm 0.0$ & $47.2 \pm 1.8$ & $45.2 \pm 2.2$ & $50.3 \pm 2.6$ & $45.0 \pm 1.2$ & $46.8$ \\
    DANN & $40.1 \pm 0.9$ & $42.3 \pm 0.1$ & $35.9 \pm 2.1$ & $43.2 \pm 2.1 $ & $49.6 \pm 1.0$ & $47.9 \pm 1.2$ & $48.7 \pm 2.0$ & $43.9 $\\
    VREx & $39.8 \pm 1.0$ & $35.3 \pm 1.2$ & $45.7 \pm 2.1$ & $46.4 \pm 0.7 $ & $47.6 \pm 2.0$ & $54.3 \pm 1.8$ & $49.1 \pm 2.3$ & $45.5 $\\
    SRGNN & $40.8 \pm 1.0$ & $52.7 \pm 3.0$ & $42.7 \pm 0.6$ & $48.7 \pm 0.1$ & $35.2\pm 2.4$ & $56.0 \pm 2.3$ & $45.0 \pm0.9$ & $45.9 $\\
    Mixup & $38.4\pm 0.9$ & $47.9 \pm 0.4$ & $47.1 \pm 2.0$ & $38.7 \pm 1.0$ & $45.5\pm 2.0$ & $57.6 \pm 0.6$ & \textbf{53.1} $\pm$ \textbf{2.1} & $46.9 $\\
    GroupDRO &\underline{$46.0 \pm 0.1$} & $53.0\pm 1.9$ & $50.7 \pm 2.0$ & $52.2 \pm 0.9$ & $51.3\pm 2.8$ & $60.4 \pm 1.8$ & $46.9 \pm 0.7$ & $51.5 $\\
   \midrule
   \sysname{}-C &  $35.1 \pm 1.2$ & $58.5\pm 1.1$ & $50.0 \pm 1.0$ & $54.2 \pm 1.0$ & $56.6\pm 1.9$ & $58.4 \pm 1.3$ & $49.0 \pm 0.6$ & $51.7$ \\
   \sysname{}-A & $35.7 \pm 0.9$ & $59.1 \pm 1.5$ & $50.2 \pm 0.8$ & $54.0 \pm 1.3$ & \underline{$59.7\pm 2.2$ }& $62.9\pm 1.7$ & $48.8\pm 1.5$ & \underline{$52.9$}\\
   \sysname{}& \textbf{46.0} $\pm$ \textbf{3.3} & \textbf{61.2}$ \pm$ \textbf{2.6} & \underline{$51.9 \pm 2.2$} & \underline{$55.1 \pm 2.2$} & \textbf{60.0} $\pm$ \textbf{3.6} & \underline{$64.9 \pm 3.9$} & \underline{$49.3 \pm 2.1$ } &  $\textbf{55.0}$ \\  
    \bottomrule    
  \end{tabular}
\end{table*}

\begin{table*}[t]
  \footnotesize
  \setlength\tabcolsep{2pt}
  \caption{F1 score (\%) on \textsc{Fb-100} where we compare different configurations of training graphs.}
  \label{tab.f1_fb100}
  \centering
  \begin{tabular}{lccc|c | ccc|c | ccc | c}
    \toprule
    ~ & \multicolumn{4}{c}{\textbf{Hopkins + Caltech + Amherst}} & \multicolumn{4}{|c|}{\textbf{Bingham + Duke + Princeton}} & \multicolumn{4}{c}{\textbf{WashU + Brandeis+ Carnegie}}\\
    \midrule
    \textbf{Methods} & \textbf{Penn}  & \textbf{Brown} & \textbf{Texas}  & \textbf{Avg} & \textbf{Penn}  & \textbf{Brown} & \textbf{Texas}  & \textbf{Avg} & \textbf{Penn}  & \textbf{Brown} & \textbf{Texas}  & \textbf{Avg}\\
    \midrule
    EERM& $48.6 \pm 1.1$ &$47.4 \pm 3.1$ &$50.0 \pm 2.7$ & $48.6$ & $44.3 \pm 2.6$ & $39.8 \pm 7.0$ & $48.2 \pm 8.6$ & $44.1$ & $51.4 \pm 1.8$ & $46.0 \pm 2.8$ & $49.4 \pm 2.1$ & $48.9$\\
    ERM & $46.8 \pm 2.9$ & \underline{$49.1 \pm 0.0$} & \underline{$51.4 \pm 2.1$} & $49.1$ & $37.4 \pm 3.4$ & $30.2 \pm 1.4$ & $36.9 \pm 2.0$ & $34.8$ & \textbf{53.1} $\pm $ \textbf{2.8} &$43.6 \pm 4.8$ & $46.5 \pm 1.5$ & $47.7$ \\
    IRM & $44.4 \pm 0.1$ & $40.8 \pm 1.0$ & $37.4 \pm 1.2$ & $40.9$ & $32.1 \pm 0.1$ &$30.3 \pm 1.4$ & $36.9 \pm 1.0$ & $33.1$ & $39.9 \pm 1.1$ &$42.8 \pm 1.8$ & $41.4 \pm 1.8 $ & $41.4$\\
    DANN & $40.5 \pm 0.2$ & $41.4 \pm 1.2$ & $43.2 \pm 2.1$ & $41.7$ &$39.6 \pm 0.2$ & $42.8 \pm 0.8$ & $42.9 \pm 3.0$ &$41.8$ & $41.6 \pm 0.7$ & $42.9 \pm 1.2$ &$43.0 \pm 1.1$ & $42.5$\\
    VREx & $40.8 \pm 2.2$ & $48.5 \pm 1.0$ &$45.6 \pm 2.1$ & $44.7$ &$41.4 \pm 2.0$ & $46.9\pm 2.1$ & $46.8 \pm 2.8$ & $45.0$ & $42.8 \pm 0.9$ &$49.3\pm 1.0$ &$46.7 \pm 0.7$ & $46.3$\\
    SRGNN& $38.3 \pm 0.7$ & $40.0 \pm 0.9$ &$44.4 \pm 1.6$ & $40.9$ & $39.5 \pm 0.6$ &$41.8 \pm 0.7$ &$45.3 \pm 1.7$ & $42.2$ & $40.5 \pm 3.2$ & $41.1 \pm 2.6$ & $45.0 \pm 0.3$ & $42.2$\\
    Mixup& $48.2 \pm 2.0$ & $43.2 \pm 0.1 $ & $47.8 \pm 0.8$ & $46.4$& \underline{$44.9 \pm 1.0$} &$44.6 \pm 0.7$  & $48.7 \pm 0.5 $ & $46.1$ & $50.2 \pm 2.0$ & $45.5 \pm 0.1$ & $49.5 \pm 1.2$ &$48.4$\\
    GroupDRO & $46.5 \pm 0.1$ & $47.5 \pm 2.0$ &$48.3 \pm 1.0$ & $47.4$& $44.8 \pm 0.3$ & \underline{$46.9 \pm 2.0$} & $48.8 \pm 2.7$ & \underline{$46.8$ }& $49.9 \pm 0.4$ &$48.7\pm 2.1$ & \underline{$50.5 \pm 0.2$} & $49.7$\\  
    \midrule
   \sysname{}-C & $48.3 \pm 1.6$ & $48.7 \pm 0.6 $ & $49.6 \pm 0.2$ & $48.9$ & $44.8 \pm 2.1$ & $45.7 \pm 0.7$  & \underline{$50.0 \pm 0.3 $} & $46.8$ & $51.5 \pm 1.0$ & $48.3 \pm 0.7$ & $50.2 \pm 1.3$ &$50.0$ \\
   \sysname{}-A & \underline{$48.9 \pm 1.0$} & $49.1 \pm 0.7 $ & $50.9 \pm 1.3$ & \underline{$49.6$} & $44.1 \pm 2.0$ & $46.3 \pm 0.5$  & $49.8 \pm 1.7 $ & $46.7$ & $51.2 \pm 1.3$ & \underline{$50.1 \pm 0.9$} & $50.5\pm 1.6$ & \underline{$50.6$}\\
   \sysname{} & \textbf{49.4} $\pm$ \textbf{2.0} & \textbf{49.8} $\pm$ \textbf{1.5} & \textbf{51.8} $\pm$ \textbf{2.8} & \textbf{51.3}& \textbf{45.0} $\pm$ \textbf{2.1} & \textbf{47.5} $\pm$ \textbf{1.0} & \textbf{51.9} $\pm$ \textbf{1.9} & \textbf{48.1} & \underline{$52.9 \pm 2.1$} & \textbf{50.8} $\pm$ \textbf{1.9} & \textbf{51.0} $\pm$ \textbf{0.8}& \textbf{51.6}\\ 
    \bottomrule
  \end{tabular}
\end{table*}

% \clearpage

\end{document}